\documentclass{article}
\usepackage{neurips_2022}
\usepackage[utf8]{inputenc} 
\usepackage[T1]{fontenc}    
\usepackage{hyperref}       
\hypersetup{
    colorlinks=true,
    linkcolor=blue,
    filecolor=magenta,      
    urlcolor=cyan,
    citecolor=blue,
    pdftitle={Overleaf Example},
    pdfpagemode=FullScreen,
    }
    
\usepackage{url}            
\usepackage{booktabs}       
\usepackage{amsfonts}       
\usepackage{nicefrac}       
\usepackage{microtype}      
\usepackage{xcolor}         
\usepackage{algorithmic}
\usepackage{graphicx}
\usepackage{algorithm}
\usepackage{amsmath}
\usepackage{float}
\usepackage{wrapfig}
\usepackage{tikz}

\newcommand{\tom}[1]{{\color{orange}Tom: #1}}

\newcommand{\reals}{\mathbb{R}}

\usepackage[normalem]{ulem}

\usepackage{soul}

\title{Cold Diffusion: Inverting Arbitrary Image Transforms Without Noise}

\author{
\normalsize \textbf{Arpit Bansal$^1$ 
\quad Eitan Borgnia$^{*1}$ 
\quad Hong-Min Chu$^{*1}$
\quad Jie S. Li$^{1}$
} \\
\normalsize  \textbf{Hamid Kazemi$^1$ 
\quad Furong Huang$^1$ 
\quad Micah Goldblum$^{2}$} \\  
\normalsize  \textbf{Jonas Geiping$^1$ 
\quad Tom Goldstein$^1$}\\ 
\normalsize $^1$University of Maryland \quad$^2$New York University 
}

\begin{document}

\maketitle

\begin{abstract} 
Standard diffusion models involve an image transform  -- adding Gaussian noise -- and an image restoration operator that inverts this degradation.  We observe that the generative behavior of diffusion models is not strongly dependent on the choice of image degradation, and in fact an entire family of generative models can be constructed by varying this choice. Even when using completely deterministic degradations (e.g., blur, masking, and more), the training and test-time update rules that underlie diffusion models can be easily generalized to create generative models. 
%
%
The success of these fully deterministic models calls into question the community's understanding of diffusion models, which relies on noise in either gradient Langevin dynamics or variational inference, and paves the way for generalized diffusion models that invert arbitrary processes.  Our code is available at \href{https://github.com/arpitbansal297/Cold-Diffusion-Models}{github.com/arpitbansal297/Cold-Diffusion-Models}.
\end{abstract}

\begin{figure}[bh]
    \flushright
\,\,\, Original$\xrightarrow{\text{\hspace{1.8cm}Forward\hspace{1.8cm}}}$
Degraded
$\xrightarrow{\text{\hspace{1.8cm}Reverse\hspace{1.8cm}}}$
Generated
    \begin{tikzpicture}
    \hspace{0mm}
    \node (img1){
     \includegraphics[width=.98\textwidth]{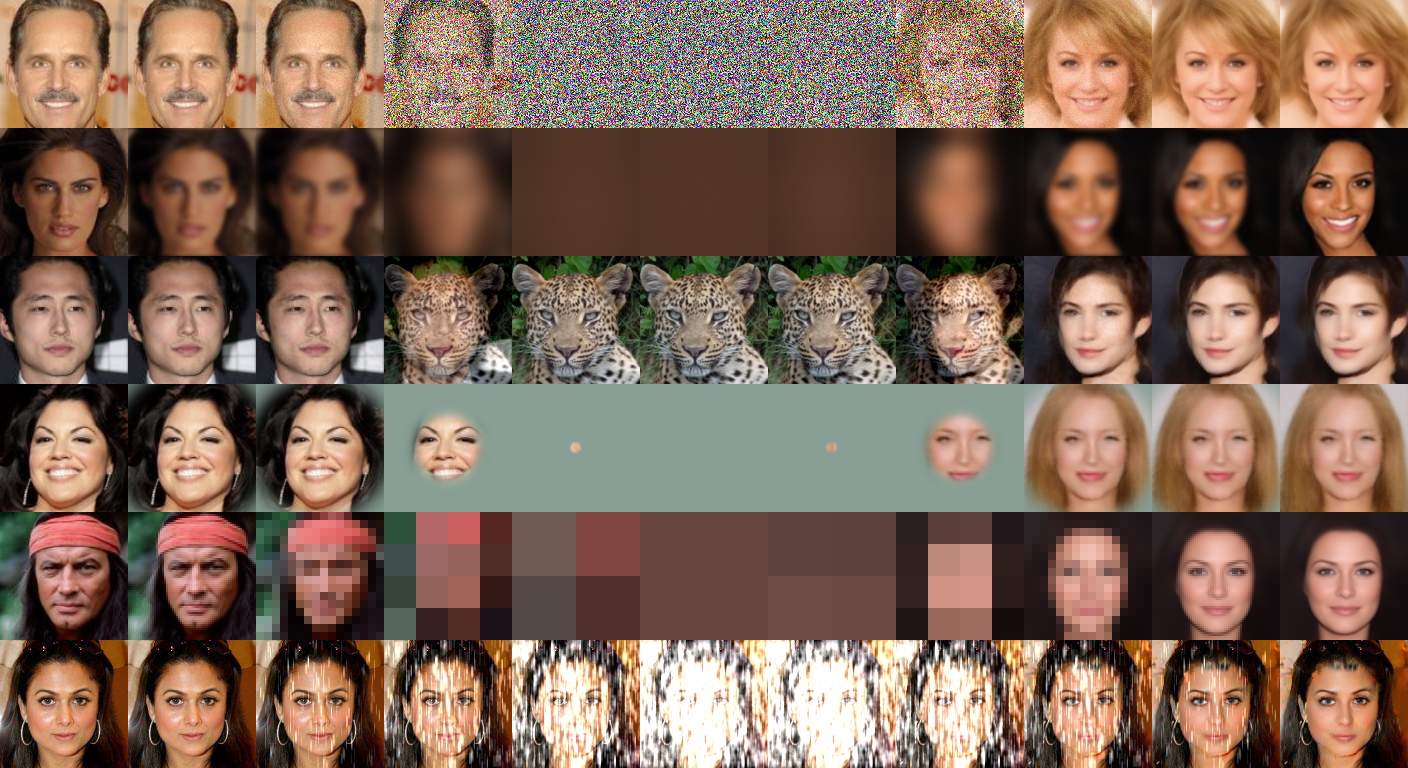}};
    \hspace{-2mm} \node[left of=img1, node distance=0cm, rotate=90, yshift=6.9cm]{ Snow \,\, Pixelate \,\, Mask  Animorph \, Blur \,\,\,\,\, Noise\,\,};
     \end{tikzpicture}\\
     \vspace{-2mm}
    \caption{Demonstration of the forward and backward processes for both hot and cold diffusions. 
    While standard diffusions are built on Gaussian noise (top row), we show that generative models can be built on arbitrary and even noiseless/cold image transforms, including the ImageNet-C \textit{snowification} operator, and an \emph{animorphosis} operator that adds a random animal image from AFHQ.  
    }
    \label{fig:all_transforms_cover}
\end{figure}

\section{Introduction}

Diffusion models have recently emerged as powerful tools for generative modeling \citep{ramesh2022hierarchical}. Diffusion models come in many flavors, but all are built around the concept of random noise removal; one trains an image restoration/denoising network that accepts an image contaminated with Gaussian noise, and outputs a denoised image. 
At test time, the denoising network is used to convert pure Gaussian noise into a photo-realistic image using an update rule that alternates between applying the denoiser and adding Gaussian noise.  When the right sequence of updates is applied, complex generative behavior is observed.

The origins of diffusion models, and also our theoretical understanding of these models, are strongly based on the role played by Gaussian noise during training and generation. Diffusion has been understood as a random walk around the image density function using {\em Langevin dynamics} \citep{sohl2015thermaldiffusion,song2019generative}, which requires Gaussian noise in each step.  The walk begins in a high temperature (heavy noise) state, and slowly anneals into a ``cold'' state with little if any noise.  Another line of work derives the loss for the denoising network using variational inference with a Gaussian prior \citep{DDPM_Ho2020,DDIM_Song2021,Nichol2021improved_ddpm}.

In this work, we examine the need for Gaussian noise, or any randomness at all, for diffusion models to work in practice. We consider {\em generalized diffusion models} that live outside the confines of the theoretical frameworks from which diffusion models arose.  Rather than limit ourselves to models built around Gaussian noise, we consider models built around arbitrary image transformations like blurring, downsampling, etc.  We train a restoration network to invert these deformations using a simple $\ell_p$ loss. When we apply a sequence of updates at test time that alternate between the image restoration model and the image degradation operation, generative behavior emerges, and we obtain photo-realistic images. 


 The existence of \textit{cold diffusions} that require no Gaussian noise (or any randomness) during training or testing raises questions about the limits of our theoretical understanding of diffusion models.  It also unlocks the door for potentially new types of generative models with very different properties than conventional diffusion seen so far.


\section{Background}

Generative models exist for a range of modalities spanning natural language~\citep{brown2020gpt3} and images \citep{brock2019gantuning,dhariwal21diffusion_beats_gan},
and they can be extended to solve important problems such as image restoration \citep{bajaht2021snips,bahjat2022ddrm}.
While GANs \citep{goodfellow2014gan} have historically been the tool of choice for image synthesis \citep{brock2019gantuning,wu2019gan_sota_exp}, diffusion models \citep{sohl2015thermaldiffusion} have recently become competitive if not superior for some applications \citep{dhariwal21diffusion_beats_gan,nichol2021glide,ramesh2021zero, SDEdit_Meng}.

Both the Langevin dynamics and variational inference interpretations of diffusion models rely on properties of the Gaussian noise used in the training and sampling pipelines. From the score-matching generative networks perspective \citep{song2019generative, scoreSDE_2021Song}, noise in the training process is critically thought to expand the support of the low-dimensional training distribution to a set of full measure in ambient space. The noise is also thought to act as data augmentation to improve score predictions in low density regions, allowing for mode mixing in the stochastic gradient Langevin dynamics (SGLD) sampling. The gradient signal in low-density regions can be further improved during sampling by injecting large magnitudes of noise in the early steps of SGLD and gradually reducing this noise in later stages.

\citet{Variational_Kingma} propose a method to learn a noise schedule that leads to faster optimization. Using a classic statistical result, ~\citet{denoiser_Kadkhodaie} show the connection between removing additive Gaussian noise and the gradient of the log of the noisy signal density in deterministic linear inverse problems. Here, we shed light on the role of noise in diffusion models through theoretical and empirical results in applications to inverse problems and image generation.

Iterative neural models have been used for various inverse problems~\citep{RED_Romano2016,ldamp_Metzler}. Recently, diffusion models have been applied to them \citep{scoreSDE_2021Song} for the problems of deblurring, denoising, super-resolution, and compressive sensing ~\citep{deblur_Whang,denoiser_Kawar,superRes_Saharia,denoiser_Kadkhodaie}.

Although not their focus, previous works on diffusion models have included experiments with deterministic image generation ~\citep{DDIM_Song2021,dhariwal21diffusion_beats_gan} and in selected inverse problems~\citep{bahjat2022ddrm}. Here, we show definitively that noise is not a necessity in diffusion models, and we observe the effects of removing noise for a number of inverse problems. 

Despite prolific work on generative models in recent years, methods to probe the properties of learned distributions and measure how closely they approximate the real training data are by no means closed fields of investigation. 
Indirect feature space similarity metrics such as Inception Score \citep{salimans2016improved}, Mode Score \citep{che2016mode}, Frechet inception distance (FID) \citep{heusel2017gans}, and Kernel inception distance (KID) \citep{binkowski2018demystifying} have been proposed and adopted to some extent, but they have notable limitations \citep{barratt2018note}. To adopt a popular frame of reference, we will use FID as the feature similarity metric for our experiments.

\section{Generalized Diffusion}
\label{sec:generalized_diffusion}
Standard diffusion models are built around two components.  First, there is an image degradation operator that contaminates images with Gaussian noise.  Second, a trained restoration operator is created to perform denoising.  The image generation process alternates between the application of these two operators. In this work, we consider the construction of generalized diffusions built around arbitrary degradation operations. These degradations can be randomized (as in the case of standard diffusion) or deterministic.

\subsection{Model components and training}
\label{sec:training}
Given an image $x_0\in \reals^N$, consider the {\em degradation} of $x_0$ by operator $D$ with severity $t,$ denoted $x_t = D(x_0,t)$.  The output distribution $D(x_0,t)$ of the degradation should vary continuously in $t,$ and the operator should satisfy
  $$D(x_0,0)=x_0.$$
In the standard diffusion framework, $D$ adds Gaussian noise with variance proportional to $t$.  In our generalized formulation, we choose $D$ to perform various other transformations such as blurring, masking out pixels, downsampling, and more, with severity that depends on $t$. We explore a range of choices for $D$ in Section~\ref{sec:restoration_exp}.
 
We also require a {\em restoration} operator $R$ that (approximately) inverts $D$.  This operator has the property that 
    $$R(x_t,t) \approx x_0.$$
In practice, this operator is implemented via a neural network parameterized by $\theta$. The restoration network is trained via the minimization problem
  \begin{equation}
  \min_\theta \mathbb{E}_{x\sim \mathcal{X}} \| R_\theta(D(x,t),t) - x \|, \label{eq:min}
  \end{equation}
where $x$ denotes a random image sampled from distribution $\mathcal{X}$ and $\|\cdot\|$ denotes a norm, which we take to be $\ell_1$ in our experiments. We have so far used the subscript $R_\theta$ to emphasize the dependence of $R$ on $\theta$ during training, but we will omit this symbol for simplicity in the discussion below.

\subsection{Sampling from the model}



After choosing a degradation $D$ and training a model $R$ to perform the restoration, these operators can be used in tandem to invert severe degradations by using standard methods borrowed from the diffusion literature. For small degradations ($t\approx 0$), a single application of $R$ can be used to obtain a restored image in one shot.  However, because $R$ is typically trained using a simple convex loss, it yields blurry results when used with large $t$. Rather, diffusion models \citep{DDIM_Song2021,DDPM_Ho2020} perform generation by iteratively applying the denoising operator and then adding noise back to the image, with the level of added noise decreasing over time. This corresponds to the standard update sequence in Algorithm \ref{alg:sa1}. 


\begin{wrapfigure}{R}{0.54\textwidth}
\begin{minipage}{0.54\textwidth}
\vspace{-10pt}
\begin{algorithm}[H]
   \caption{Naive Sampling}
   \label{alg:sa1}
    \begin{algorithmic}
    \STATE {\bfseries Input:} A degraded sample $x_t$
    \FOR{\textit{s} $=t,t-1,\dots,1$}
    \STATE $\hat x_0 \gets R(x_s, s)$ 
    \STATE $x_{s-1} = D(\hat x_0, s-1)$
    \ENDFOR
    \STATE {\bfseries Return:} $x_0$
    \end{algorithmic}
\end{algorithm}
\end{minipage}

\begin{minipage}{0.54\textwidth}
\begin{algorithm}[H]
   \caption{Improved Sampling for Cold Diffusion }
   \label{alg:sa2}
    \begin{algorithmic}
    \STATE {\bfseries Input:} A degraded sample $x_t$
    \FOR{\textit{s} $=t,t-1,\dots,1$}
    \STATE $\hat x_0 \gets R(x_s, s)$ 
    \STATE $x_{s-1} = x_s  - D(\hat x_0, s) + D(\hat x_0, s-1) $ 
    \ENDFOR
    \end{algorithmic}
\end{algorithm}
\vspace{-22pt}
\end{minipage}
\end{wrapfigure}

When the restoration operator is perfect, \textit{i.e.} when $R(D(x_0,t),t) = x_0$ for all $t,$ one can easily see that Algorithm \ref{alg:sa1} produces exact iterates of the form $x_s = D(x_0,s)$.  But what happens for imperfect restoration operators?  In this case, errors can cause the iterates $x_s$ to wander away from $D(x_0,s)$, and inaccurate reconstruction may occur.

We find that the standard sampling approach in Algorithm \ref{alg:sa1} works well for noise-based diffusion, possibly because the restoration operator $R$ has been trained to correct (random Gaussian) errors in its inputs. However, we find that it yields poor results in the case of cold diffusions with smooth/differentiable degradations as demonstrated for a deblurring model in Figure \ref{fig:algorithms}. We propose Algorithm \ref{alg:sa2} for sampling, which we find to be superior for inverting smooth, cold degradations.

This sampler has important mathematical properties that enable it to recover high quality results.  Specifically, for a class of linear degradation operations, it can be shown to produce exact reconstruction (\textit{i.e.} $x_s = D(x_0,s)$) even when the restoration operator $R$ fails to perfectly invert $D$. We discuss this in the following section.

\subsection{Properties of Algorithm \ref{alg:sa2}}
\label{sec:analysis}

\begin{wrapfigure}{R}{0.54\textwidth}
\begin{minipage}{0.54\textwidth}
\vspace{-22pt}
 \begin{figure}[H]
    \centering
    \includegraphics[width=\textwidth]{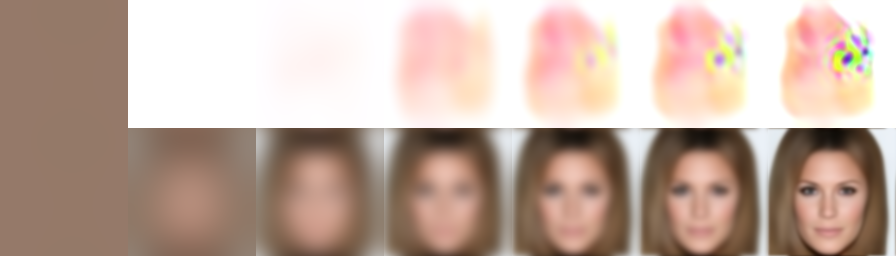} 
    \vspace{-4mm}
    \caption{Comparison of sampling methods for cold diffusion on the CelebA dataset. \textbf{Top:} Algorithm \ref{alg:sa1} produces compounding artifacts and fails to generate a new image. \textbf{Bottom:} Algorithm \ref{alg:sa2} succeeds in sampling a high quality image without noise.} 
    \vspace{-12pt}
    \label{fig:algorithms}
\end{figure}
\end{minipage}
\end{wrapfigure}

It is clear from inspection that both Algorithms \ref{alg:sa1} and \ref{alg:sa2} perfectly reconstruct the iterate $x_s = D(x_0,s)$ for all $s<t$ if the restoration operator is a perfect inverse for the degradation operator. In this section, we analyze the stability of these algorithms to errors in the restoration operator.

For small values of $x$ and $s$, Algorithm \ref{alg:sa2} is extremely tolerant of error in the restoration operator $R$.
To see why, consider a model problem with a linear degradation function of the form $D(x,s) \approx x + s\cdot e \label{firstorder}$ for some vector $e$. While this ansatz may seem rather restrictive, note that the Taylor expansion of any smooth degradation $D(x,s)$ around $x=x_0,s=0$ has the form $D(x,s) \approx x + s\cdot e + \text{HOT}$ where $\text{HOT}$ denotes higher order terms. 
Note that the constant/zeroth-order term in this Taylor expansion is zero because we assumed above that the degradation operator satisfies $D(x,0) = x$. 
 
 For a degradation of the form \eqref{firstorder} and any restoration operator $R$, the update in Algorithm \ref{alg:sa2} can be written
  \begin{align}
  x_{s-1} &= x_s - D(R(x_s,s),s) + D(R(x_s,s),s-1) \nonumber\\
   &=D(x_0,s)- D(R(x_s,s),s) + D(R(x_s,s),s-1)\nonumber \\
   &=x_0+s\cdot e - R(x_s,s) - s\cdot e + R(x_s,s) +(s-1)\cdot e\nonumber \\
   &=x_0 +(s-1)\cdot e \nonumber \\
   &= D(x_0,s-1)\nonumber 
  \end{align}

By induction, we see that the algorithm produces the value $x_s =D(x_0,s)$ for all $s<t,$ regardless of the choice of $R$.  In other words, for {\em any} choice of $R$, the iteration behaves the same as it would when $R$ is a perfect inverse for the degradation $D$.

 
 By contrast, Algorithm \ref{alg:sa1}  does not enjoy this behavior.  In fact, when $R$ is not a perfect inverse for $D$, $x_0$ is not even a fixed point of the update rule in Algorithm \ref{alg:sa1} because
  $x_0 \neq D(R(x,0),0) = R(x,0).$
If $R$ does not perfectly invert $D$ we should expect Algorithm \ref{alg:sa1} to incur errors, even for small values of $s$.  Meanwhile, for small values of $s$, the behavior of $D$ approaches its first-order Taylor expansion and Algorithm \ref{alg:sa2} becomes immune to errors in $R$. We demonstrate the stability of Algorithm \ref{alg:sa2} vs Algorithm \ref{alg:sa1} on a deblurring model in Figure \ref{fig:algorithms}.

\section{Generalized Diffusions with Various  Transformations} \label{sec:restoration_exp}

In this section, we take the first step towards cold diffusion by reversing different degradations and hence performing conditional generation. We will extend our methods to perform unconditional (i.e. from scratch) generation in Section \ref{sec:generation}. We emprically evaluate generalized diffusion models trained on different degradations with our improved sampling Algorithm \ref{alg:sa2}. We perform experiments on the vision tasks of deblurring, inpainting, super-resolution, and the unconventional task of synthetic snow removal. 
We perform our experiments on MNIST~\citep{mnist}, CIFAR-10~\citep{cifar10}, and CelebA~\citep{celeba}.
In each of these tasks, we gradually remove the information from the clean image, creating a sequence of images such that $D(x_0,t)$ retains less information than $D(x_0,t-1)$. For these different tasks, we present both qualitative and quantitative results on a held-out testing dataset and demonstrate the importance of the sampling technique described in Algorithm \ref{alg:sa2}. For all quantitative results in this section, the Frechet inception distance (FID) scores \citep{heusel2017gans} for degraded and reconstructed images are measured with respect to the testing data. Additional information about the quantitative results, convergence criteria, hyperparameters, and architecture of the models presented below can be found in the appendix.

\subsection{Deblurring}

We consider a generalized diffusion based on a Gaussian blur operation (as opposed to Gaussian noise) in which an image at step $t$ has more blur than at ${t-1}$. 
The forward process given the Gaussian kernels $\{G_s\}$ and the image $x_{t-1}$ at step $t-1$ can thus be written as
\begin{equation}
    x_t = G_t * x_{t-1} = G_t * \hdots * G_1 * x_0 = \bar{G_t} * x_0 = D(x_0,t),
\end{equation}
where $*$ denotes the convolution operator, which blurs an image using a kernel.  

\begin{figure}[H]
    \centering
    \begin{tabular}{cccc}
        Degraded & Direct & Alg. & Original \\
        \includegraphics[width=0.22\textwidth] {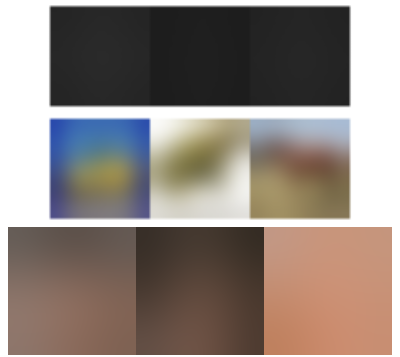} &
        \includegraphics[width=0.22\textwidth] {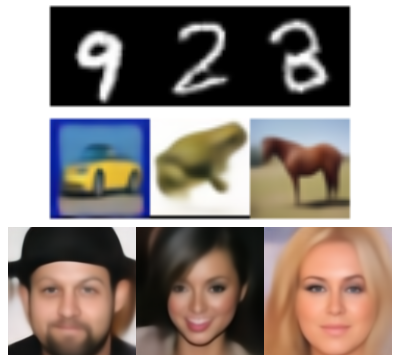} &
        \includegraphics[width=0.22\textwidth] {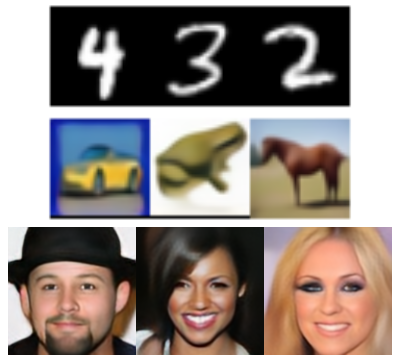} &
        \includegraphics[width=0.22\textwidth] {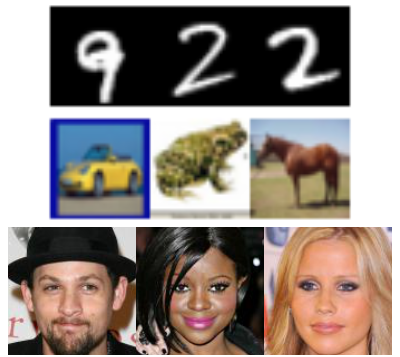}
    \end{tabular}
    \vspace{-4mm}
    \caption{Deblurring models trained on the MNIST, CIFAR-10, and CelebA datasets. \textbf{Left to right:} degraded inputs $D(x_0, T)$ , direct reconstruction $R(D(x_0, T))$, sampled reconstruction with Algorithm \ref{alg:sa2}, and original image.
    }
    \label{fig:blur_revert}
\end{figure}

We train a deblurring model by minimizing the loss \eqref{eq:min}, and then use Algorithm~\ref{alg:sa2} to invert this blurred diffusion process for which we trained a DNN to predict the clean image $\hat x_0$. Qualitative results are shown in Figure \ref{fig:blur_revert} and quantitative results in Table \ref{tab:blur_revert}. Qualitatively, we can see that images created using the sampling process are sharper and in some cases completely different as compared to the direct reconstruction of the clean image. Quantitatively we can see that the reconstruction metrics such as RMSE and PSNR get worse when we use the sampling process, but on the other hand FID with respect to held-out test data improves. The qualitative improvements and decrease in FID show the benefits of the generalized sampling routine, which brings the learned distribution closer to the true data manifold.


In the case of blur operator, the sampling routine can be thought of adding frequencies at each step. This is because the sampling routine involves the term $ D(\hat{x_0}, t) - D(\hat{x_0}, t-1)$ which in the case of blur becomes $\bar{G}_{t}*x_0 - \bar{G}_{t-1}*x_0$. This results in a difference of Gaussians, which is a band pass filter and contains frequencies that were removed at step $t$. Thus, in the sampling process, we sequentially add the frequencies that were removed during the degradation process.

\begin{table}[h]
\centering
\footnotesize
\caption{Quantitative metrics for quality of image reconstruction using deblurring models.}
\vspace{-3mm} 
\label{tab:blur_revert}
\vspace{0.1in}
\begin{tabular}{c|ccc|ccc|ccc}
\toprule
 & & \textbf{Degraded} & & & \textbf{Sampled} & & & \textbf{Direct}\\
Dataset & FID &  SSIM &  RMSE & FID &  SSIM &  RMSE & FID & SSIM & RMSE\\
\midrule
MNIST & 438.59 & 0.287 & 0.287 & \textbf{4.69} & 0.718 & 0.154 & 5.10 & \textbf{0.757} & 0.142 \\
CIFAR-10 & 298.60 & 0.315 & 0.136 & \textbf{80.08} & 0.773 & 0.075 & 83.69 & \textbf{0.775} & 0.071 \\
CelebA & 382.81 & 0.254 & 0.193 & \textbf{26.14} & 0.568 & 0.093 & 36.37 & \textbf{0.607} & 0.083 \\
\bottomrule
\end{tabular}
\end{table}
  
\subsection{Inpainting}
\label{sec:inpainting}

\begin{figure}[b]
    \centering
    \begin{tabular}{cccc}
        Degraded & Direct & Alg. & Original \\
        \includegraphics[width=0.22\textwidth] {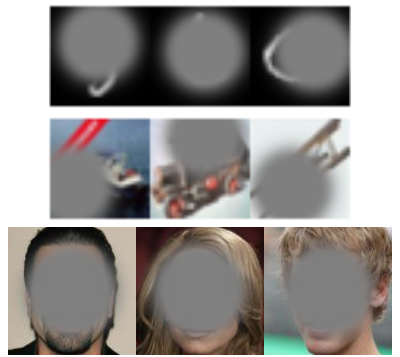} &
        \includegraphics[width=0.22\textwidth] {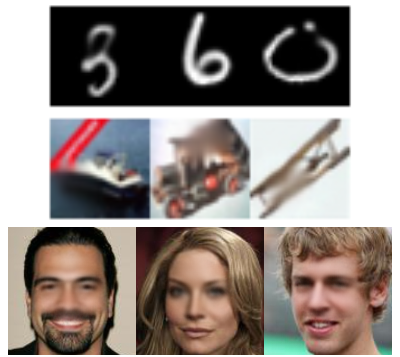} &
        \includegraphics[width=0.22\textwidth] {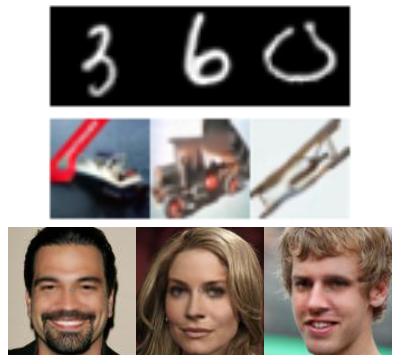} &
        \includegraphics[width=0.22\textwidth] {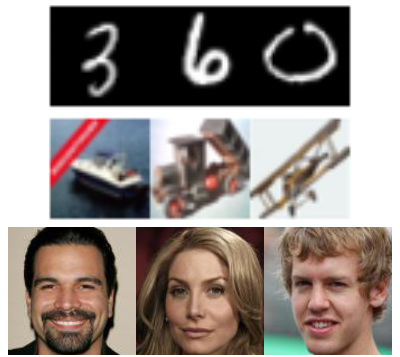}
    \end{tabular}
    \vspace{-4mm}
    \caption{Inpainting models trained on the MNIST, CIFAR-10, and CelebA datasets. \textbf{Left to right:} Degraded inputs $D(x_0, T)$ , direct reconstruction $R(D(x_0, T))$, sampled reconstruction with Algorithm \ref{alg:sa2}, and original image.
    }
    \label{fig:fade_revert}
\end{figure}

We define a schedule of transforms that progressively grays-out pixels from the input image. We remove pixels using a Gaussian mask as follows: For input images of size $n \times n$ we start with a 2D Gaussian curve of variance $\beta,$ discretized into an $n \times n$ array. We normalize so the peak of the curve has value 1, and subtract the result from 1 so the center of the mask as value 0.  We randomize the location of the Gaussian mask for MNIST and CIFAR-10, but keep it centered for CelebA. We denote the final mask by $z_{\beta}$. 

Input images $x_0$ are iteratively masked for $T$ steps via multiplication with a sequence of masks $\{z_{\beta_i}\}$ with increasing $\beta_i$. We can control the amount of information removed at each step by tuning the $\beta_i$ parameter. In the language of Section \ref{sec:generalized_diffusion}, $D(x_0, t) = x_0  \cdot \prod_{i=1}^{t} z_{\beta_i}$, where the operator $\cdot$ denotes entry-wise multiplication. 

Figure \ref{fig:fade_revert} presents results on test images and compares the output of the inpainting model to the original image. The reconstructed images display reconstructed features qualitatively consistent with the context provided by the unperturbed regions of the image. We quantitatively assess the effectiveness of the inpainting models on each of the datasets by comparing distributional similarity metrics before and after the reconstruction. Our results are summarized in Table \ref{tab:fade_revert}.  Note, the FID scores here are computed with respect to the held-out validation set. 

\begin{table}[h]
\centering
\footnotesize
\caption{Quantitative metrics for quality of image reconstruction using inpainting models.}
\label{tab:fade_revert}
\vspace{-1mm}
\begin{tabular}{c|ccc|ccc|ccc}
\toprule
 & & \textbf{Degraded} & & & \textbf{Sampled} & & & \textbf{Direct}\\
Dataset & FID &  SSIM &  RMSE & FID &  SSIM &  RMSE & FID & SSIM & RMSE\\
\midrule
MNIST              & 108.48 & 0.490 & 0.262 & \textbf{1.61} & 0.941 & 0.068 & 2.24 & \textbf{0.94}8 & 0.060 \\
CIFAR-10           & 40.83 & 0.615 & 0.143 & \textbf{8.92} & 0.859 & 0.068 & 9.97 & \textbf{0.869} & 0.063 \\
CelebA             & 127.85 & 0.663 & 0.155 & \textbf{5.73} & 0.917 &  0.043 & 7.74 & \textbf{0.922} & 0.039 \\
\bottomrule
\end{tabular}
\end{table}

\subsection{Super-Resolution}

\begin{figure}[h]
    \centering
    \begin{tabular}{cccc}
        Degraded & Direct & Alg. & Original \\
        \includegraphics[width=0.22\textwidth] {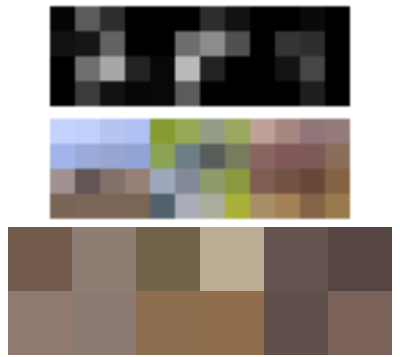} &
        \includegraphics[width=0.22\textwidth] {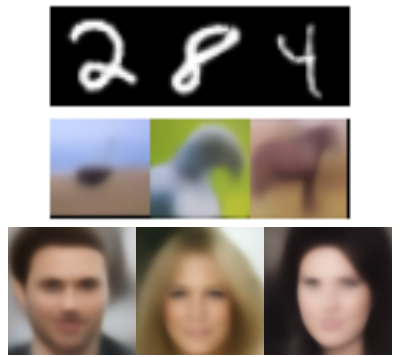} &
        \includegraphics[width=0.22\textwidth] {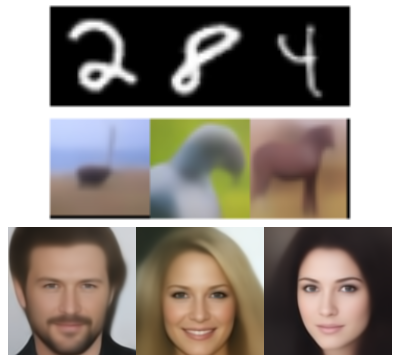} &
        \includegraphics[width=0.22\textwidth] {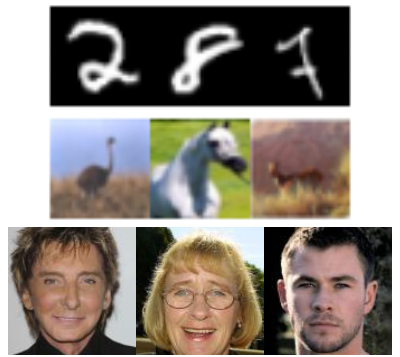}
    \end{tabular}
    \vspace{-4mm}
    \caption{Superresolution models trained on the MNIST, CIFAR-10, and CelebA datasets. \textbf{Left to right:} degraded inputs $D(x_0, T)$ , direct reconstruction $R(D(x_0, T))$, sampled reconstruction with Algorithm \ref{alg:sa2}, and original image.
    }
    \label{fig:downsample_revert}
\end{figure}

For this task, the degradation operator downsamples the image by a factor of two in each direction.  This takes place, once for each values of $t$, until a final resolution is reached, 4$\times$4 in the case of MNIST and CIFAR-10 and 2$\times$2 in the case of Celeb-A. After each down-sampling, the lower-resolution image is resized to the original image size, using nearest-neighbor interpolation. Figure \ref{fig:downsample_revert} presents example testing data inputs for all datasets and compares the output of the super-resolution model to the original image. Though the reconstructed images are not perfect for the more challenging datasets, the reconstructed features are qualitatively consistent with the context provided by the low resolution image.

Table \ref{tab:res_revert} compares the distributional similarity metrics between degraded/reconstructed images and test samples. 

\begin{table}[h]
\centering
\footnotesize
\caption{Quantitative metrics for quality of image reconstruction using super-resolution models.}
\label{tab:res_revert}
\vspace{-1mm}
\begin{tabular}{c|ccc|ccc|ccc}
\toprule
 & & \textbf{Degraded} & & & \textbf{Sampled} & & & \textbf{Direct}\\
Dataset & FID &  SSIM &  RMSE & FID &  SSIM &  RMSE & FID & SSIM & RMSE\\
\midrule
MNIST              & 368.56 & 0.178 & 0.231 & 4.33 & 0.820 & 0.115 & \textbf{4.05} & \textbf{0.823} & 0.114 \\
CIFAR-10           & 358.99 & 0.279 & 0.146 & \textbf{152.76 }& 0.411 & 0.155 & 169.94 & \textbf{0.420} & 0.152 \\
CelebA             & 349.85 & 0.335 & 0.225 & \textbf{96.92} & 0.381 &  0.201 & 112.84 & \textbf{0.400} & 0.196 \\
\bottomrule
\end{tabular}
\end{table}

 \subsection{Snowification}

Apart from traditional degradations, we additionally provide results for the task of synthetic snow removal using the \href{https://github.com/hendrycks/robustness}{offical implementation} of the \emph{snowification} transform from ImageNet-C~\citep{HendrycksD19imagenetc}. The purpose of this experiment is to demonstrate that generalized diffusion can succeed even with exotic transforms that lack the scale-space and compositional properties of blur operators.  Similar to other tasks, we degrade the images by adding \emph{snow}, such that the level of \emph{snow} increases with step $t$. We provide more implementation details in Appendix.  

\begin{figure}[H]
    \centering
    \begin{tabular}{cccc}
        Degraded & Direct & Alg. & Original \\
        \includegraphics[width=0.22\textwidth] {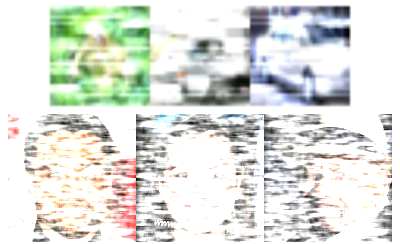} &
        \includegraphics[width=0.22\textwidth] {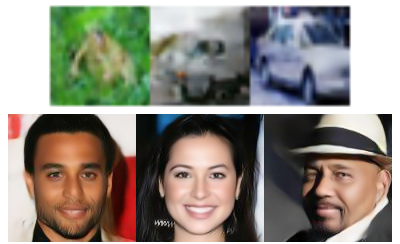} &
        \includegraphics[width=0.22\textwidth] {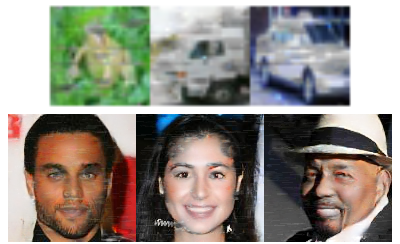} &
        \includegraphics[width=0.22\textwidth] {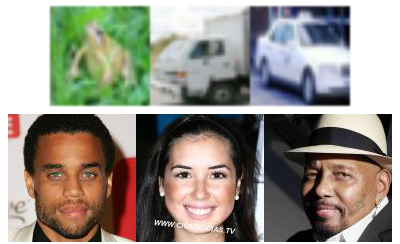}
    \end{tabular}
    \vspace{-4mm}
    \caption{\emph{Desnowification} models trained on the CIFAR-10, and CelebA datasets. \textbf{Left to right:} degraded inputs $D(x_0, T)$ , direct reconstruction $R(D(x_0, T))$, sampled reconstruction with Algorithm \ref{alg:sa2}, and original image.
    }
    \label{fig:snow_revert}
\end{figure}

We illustrate our \textit{desnowification} results in Figure \ref{fig:snow_revert}. We present testing examples, as well as their snowified images, from all the datasets, and compare the desnowified results with the original images. The desnowified images feature near-perfect reconstruction results for CIFAR-10 examples with lighter snow, and exhibit visually distinctive restoration for Celeb-A examples with heavy snow. We provide quantitative results in Table~\ref{tab:desnowify_revert}.

\begin{table}[t]
\centering
\footnotesize
\caption{Quantitative metrics for quality of image reconstruction using \textit{desnowification} models.}
\label{tab:desnowify_revert}
\vspace{-1mm}
\begin{tabular}{c|ccc|ccc}
\toprule
 & & \textbf{Degraded Image} & & & \textbf{Reconstruction} &\\
Dataset & FID &  SSIM &  RMSE & FID &  SSIM &  RMSE\\
\midrule
CIFAR-10           & 125.63 & 0.419 & 0.327 & 31.10 & 0.074 & 0.838 \\
CelebA             & 398.31 & 0.338 & 0.283 & 27.09 & 0.033 & 0.907 \\ 
\bottomrule
\end{tabular}
\end{table}

\section{Cold Generation}
\label{sec:generation}
Diffusion models can successfully learn the underlying distribution of training data, and thus generate diverse, high quality images \citep{DDIM_Song2021, dhariwal21diffusion_beats_gan, Jolicoeur-Martineau21_adv, Cascade_22Ho}. We will first discuss deterministic generation using Gaussian noise and then discuss in detail unconditional generation using deblurring. Finally, we provide a proof of concept that the Algorithm \ref{alg:sa2} can be extended to other degradations.

\subsection{Generation using deterministic noise degradation}
Here we discuss image generation using noise-based degradation.  We consider ``deterministic'' sampling in which the noise pattern is selected and frozen at the start of the generation process, and then treated as a constant. 
We study two ways of applying Algorithm \ref{alg:sa2} with fixed noise. We first define 
\begin{equation*}
    D(x, t) =  \sqrt{\alpha_t}x + \sqrt{1-\alpha_t}z,
\end{equation*}
as the (deterministic) interpolation between data point $x$ and a fixed noise pattern $z \in \mathcal{N}(0, 1)$, for increasing $\alpha_t < \alpha_{t-1}$,\, $\forall$  $1 \le t \le T$ as in \citet{DDIM_Song2021}.
 Algorithm \ref{alg:sa2} can be applied in this case by fixing the noise $z$ used in the degradation operator $D(x, s)$. Alternatively, one can deterministically calculate the noise vector $z$ to be used in step $t$ of reconstruction by using the formula
\begin{equation*}
    \hat z(x_t, t) =  \frac{x_t - \sqrt{\alpha_t} R(x_t, t)}{\sqrt{1-\alpha_t}}.
\end{equation*}
The second method turns out to be closely related to the deterministic sampling proposed in \citet{DDIM_Song2021}, with some differences in the formulation of the training objective. 
We discuss this relationship in detail in Appendix \ref{appendix:app_gen_noise}. We present quantitative results for CelebA and AFHQ datasets using the fixed noise method and the estimated noise method (using $\hat{z}$) in Table \ref{tab:fid_gen_method_2}.

\subsection{Image generation using blur}
The forward diffusion process in noise-based diffusion models has the advantage that the degraded image distribution at the final step $T$ is simply an isotropic Gaussian. One can therefore perform (unconditional) generation by first drawing a sample from the isotropic Gaussian, and sequentially denoising it with backward diffusion. 

When using blur as a degradation, the fully degraded images do not form a nice closed-form distribution that we can sample from.   They do, however, form a simple enough distribution that can be modeled with simple methods.  Note that every image $x_0$ degenerates to an $x_T$ that is constant (i.e., every pixel is the same color) for large $T$. Furthermore, the constant value is exactly the channel-wise mean of the RGB image $x_0$, and can be represented with a 3-vector. 
This 3-dimensional distribution is easily represented using a Gaussian mixture model (GMM).  This GMM can be sampled to produce the random pixel values of a severely blurred image, which can be deblurred using cold diffusion to create a new image.

Our generative model uses a blurring schedule 
where we progressively blur each image with a Gaussian kernel of size 27x27 over 300 steps.  The standard deviation of the kernel starts at 1 and increases exponentially at the rate of 0.01. We then fit a simple GMM with one component to the distribution of channel-wise means. To generate an image from scratch, we sample the channel-wise mean from the GMM, expand the 3D vector into a $128 \times 128$ image with three channels, and then apply Algorithm \ref{alg:sa2}.

Empirically, the presented pipeline generates images with high fidelity but low diversity, as reflected quantitatively by comparing the perfect symmetry column with results from hot diffusion in Table~\ref{tab:fid_gen_method_2}. We attribute this to the perfect correlation between pixels of $x_T$ sampled from the channel-wise mean Gaussian mixture model. To break the symmetry between pixels, we add 
a small amount of Gaussian noise (of standard deviation $0.002$) to each sampled $x_T$.
As shown in Table~\ref{tab:fid_gen_method_2}, the simple trick drastically improves the quality of generated images. 
We also present the qualitative results for cold diffusion using blur transformation in Figure \ref{fig:gen_method_2}, and further discuss the necessity of Algorithm \ref{alg:sa2} for generation in Appendix \ref{appendix:a7}. \\

\begin{table}[h]
\centering
\footnotesize
\caption{FID scores for CelebA and AFHQ datasets using hot (using noise) and cold diffusion (using blur transformation). This table shows that This table also shows that breaking the symmetry withing pixels of the same channel further improves the FID scores.}
\label{tab:fid_gen_method_2}
\vspace{0.1in}
\begin{tabular}{c|cc|cc}
\toprule
& \multicolumn{2}{c}{\textbf{Hot Diffusion}} & \multicolumn{2}{c}{\textbf{Cold Diffusion}} \\
Dataset         & Fixed Noise & Estimated Noise &  Perfect symmetry & Broken symmetry\\
\midrule
CelebA          & 59.91 & 23.11 & 97.00 & 49.45 \\
AFHQ            & 25.62 & 20.59 & 93.05 & 54.68 \\ 
\bottomrule
\end{tabular}
\end{table}

\begin{figure}[h]
    \centering
    \includegraphics[width=\textwidth]{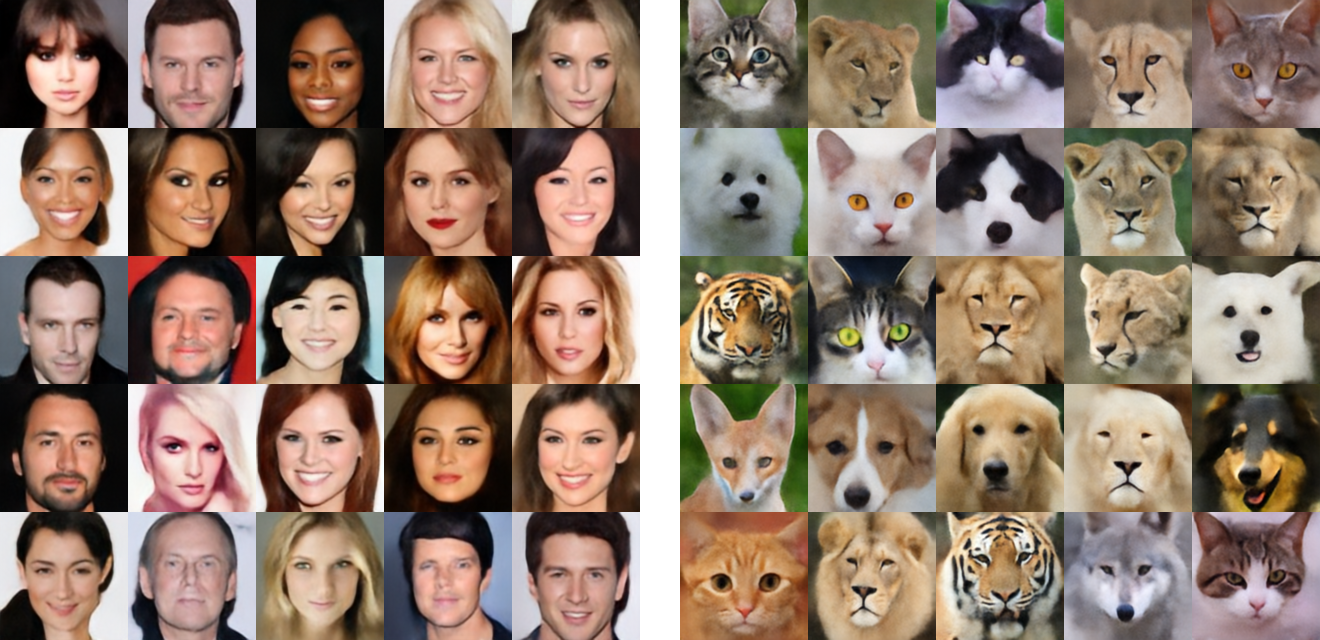}
    \caption{Examples of generated samples from $128 \times 128$ CelebA and AFHQ datasets using cold diffusion with blur transformation}
    \label{fig:gen_method_2}
    \vspace{-8pt}
\end{figure}

\subsection{Generation using other transformations}
In this section, we further provide a proof of concept that generation can be extended to other transformations. Specifically, we show preliminary results on inpainting, super-resolution, and \emph{animorphosis}. Inspired by the simplicity of the degraded image distribution for the blurring routine presented in the previous section, we use degradation routines with predictable final distributions here as well. \\

To use the Gaussian mask transformation for generation, we modify the masking routine so the final degraded image is completely devoid of information. One might think a natural option is to send all of the images to a completely black image $x_T$, but this would not allow for any diversity in generation. To get around this maximally non-injective property, we instead make the mask turn all pixels to a random, solid color. This still removes all of the information from the image, but it allows us to recover different samples from the learned distribution via Algorithm \ref{alg:sa2} by starting off with different color images. More formally, a Gaussian mask $G_t = \prod_{i=1}^{t} z_{\beta_i}$ is created in a similar way as discussed in the Section \ref{sec:inpainting}, but instead of multiplying it directly to the image $x_{0}$, we create $x_t$ as follows:
$$x_{t} = G_{t} * x_{0} + (1-G_{t}) * c$$
where $c$ is an image of a randomly sampled color.

For super-resolution, the routine down-samples to a resolution of $2 \times 2$, or $4$ values in each channel. These degraded images can be represented as one-dimensional vectors, and their distribution is modeled using one Gaussian distribution. Using the same methods described for generation using blurring described above, we sample from this Gaussian-fitted distribution of the lower-dimensional degraded image space and pass this sampled point through the generation process trained on super-resolution data to create one output.

Additionally to show one can invert nearly any transformation, we include a new transformation deemed \emph{animorphosis}, where we iteratively transform a human face from CelebA to an animal face from AFHQ. Though we chose CelebA and AFHQ for our experimentation, in principle such interpolation can be done for any two initial data distributions. 

More formally, given an image $x$ and a random image $z$ sampled from the AFHQ manifold, $x_t$ can be written as follows:
$$x_t = \sqrt{\alpha_t}x + \sqrt{1-\alpha_t}z$$
Note this is essentially the same as the noising procedure, but instead of adding noise we are adding a progressively higher weighted AFHQ image. In order to sample from the learned distribution, we sample a random image of an animal and use Algorithm \ref{alg:sa2} to reverse the \emph{animorphosis} transformation.

\begin{figure}[h]
    \centering
    \includegraphics[width=0.8\textwidth]{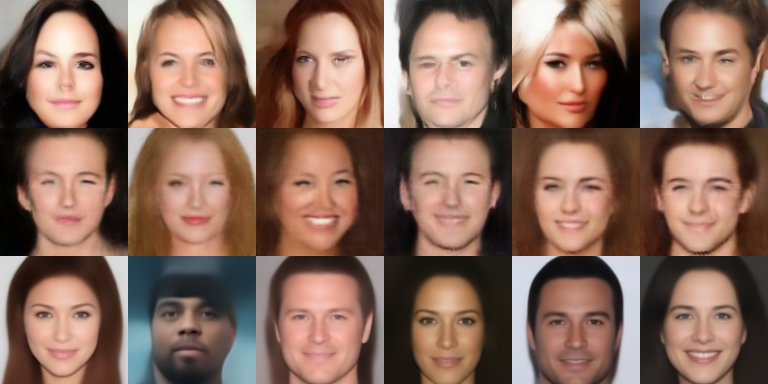}
    \caption{Preliminary demonstration of the generative abilities of other cold diffusins on the $128 \times 128$ CelebA dataset. The top row is with \emph{animorphosis} models, the middle row is with inpainting models, and the bottom row exhibits super-resolution models.}
    \label{fig:other_cold_diffusion}
\end{figure}

We present results for the CelebA dataset, and hence the quantitative results in terms of FID scores for inpainting, super-resolution and \emph{animorphosis} are 90.14, 92.91 and 48.51 respectively. We further show some qualitative samples in Figure \ref{fig:other_cold_diffusion}, and in Figure \ref{fig:all_transforms_cover}.

\section{Conclusion}
Existing diffusion models rely on Gaussian noise for both forward and reverse processes.  In this work, we find that the random noise can be removed entirely from the diffusion model framework, and replaced with arbitrary transforms.  In doing so, our generalization of diffusion models and their sampling procedures allows us to restore images afflicted by deterministic degradations such as blur, inpainting and downsampling. This framework paves the way for a more diverse landscape of diffusion models beyond the Gaussian noise paradigm.  The different properties of these diffusions may prove useful for a range of applications, including image generation and beyond.

 \label{sec:limitations}
\bibliography{ref}
\bibliographystyle{plainnat} 

\newpage
\appendix
\section{Appendix}

\subsection{Deblurring}
For the deblurring experiments, we train the models on different datasets for 700,000 gradient steps. We use the Adam \citep{kingma2014adam} optimizer with learning rate $2\times10^{-5}$. The training was done on the batch size of 32, and we accumulate the gradients every 2 steps. Our final model is an Exponential Moving Average of the trained model with decay rate 0.995 which is updated after every 10 gradient steps. \\
For the MNIST dataset, we blur recursively 40 times, with a discrete Gaussian kernel of size 11x11 and a standard deviation 7. In the case of CIFAR-10, we recursively blur with a Gaussian kernel of fixed size 11x11, but at each step $t$, the standard deviation of the Gaussian kernel is given by $0.01*t + 0.35$. The blur routine for CelebA dataset involves blurring images with a Gaussian kernel of 15x15 and the standard deviation of the Gaussian kernel grows exponentially with time $t$ at the rate of $0.01$.

\begin{figure}[h]
    \centering
    \begin{tabular}{cccc}
        Degraded & Direct & Alg. & Original \\
        \includegraphics[width=0.22\textwidth] {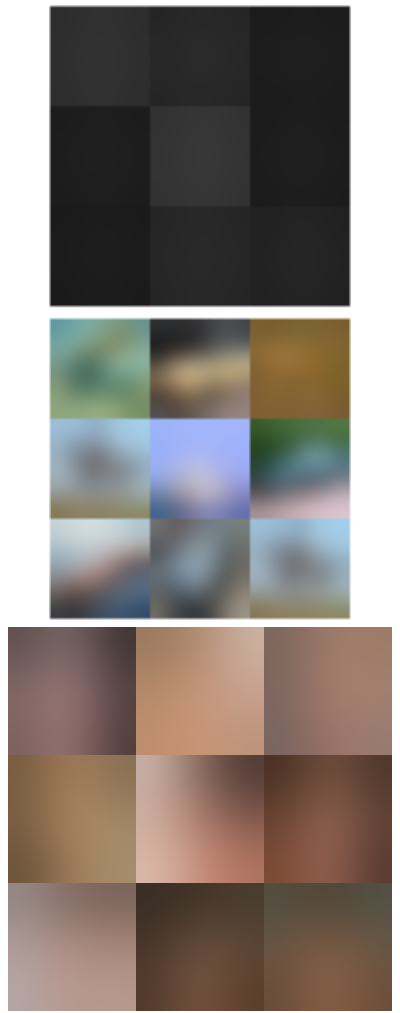} &
        \includegraphics[width=0.22\textwidth] {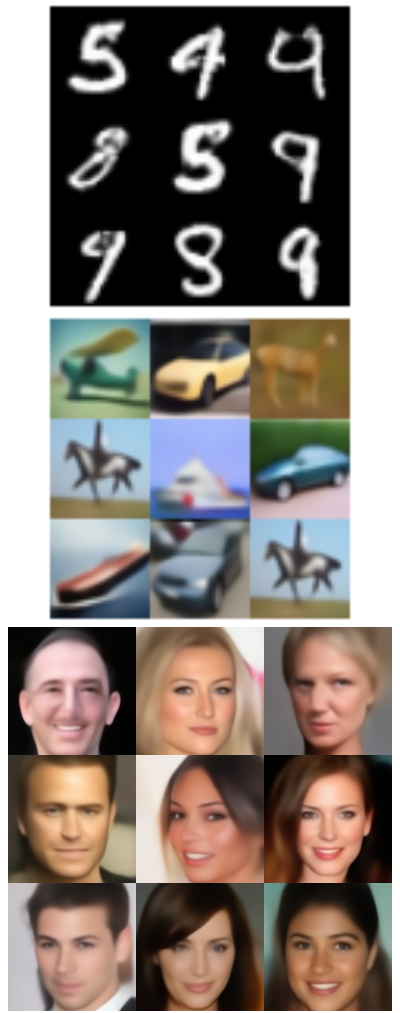} &
        \includegraphics[width=0.22\textwidth] {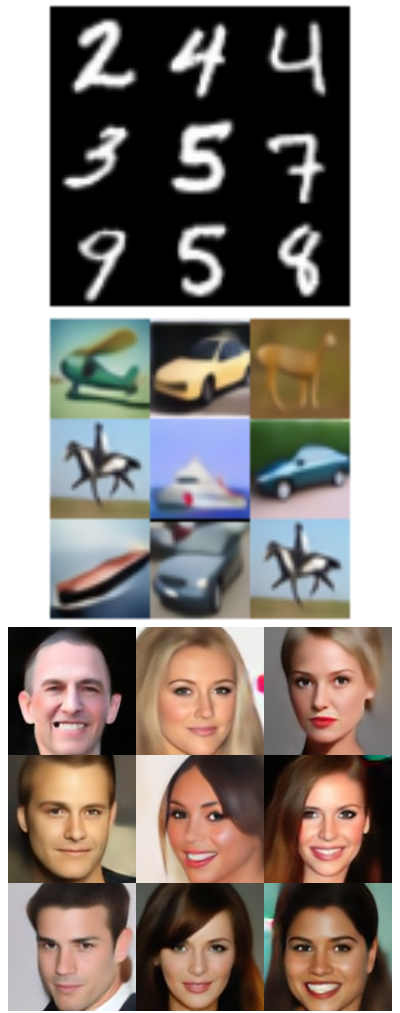} &
        \includegraphics[width=0.22\textwidth] {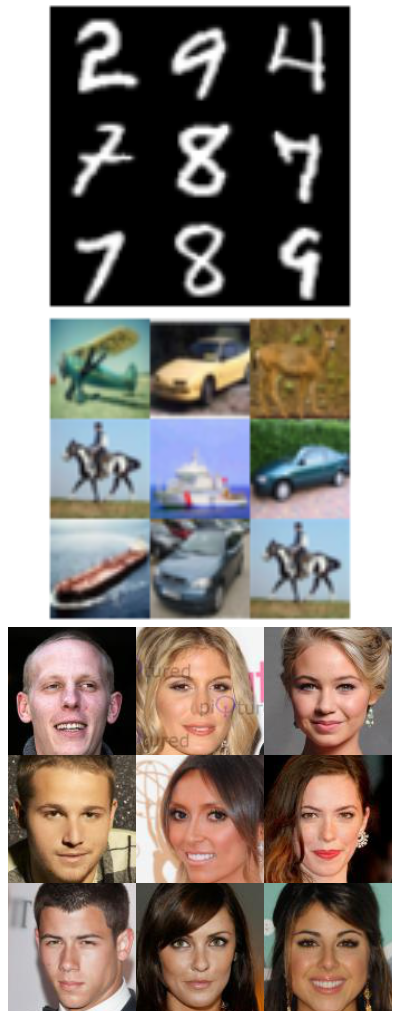}
    \end{tabular}
    \vspace{-4mm}
    \caption{Additional examples from deblurring models trained on the MNIST, CIFAR-10, and CelebA datasets. \textbf{Left to right:} degraded inputs $D(x_0, T)$ , direct reconstruction $R(D(x_0, T))$, sampled reconstruction with Algorithm \ref{alg:sa2}, and original image.
    }
    \label{fig:blur_revert_app}
\end{figure}

\begin{figure}
    \centering
    \includegraphics[width=\textwidth]{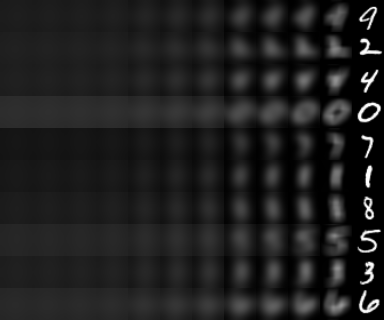}
    \includegraphics[width=\textwidth]{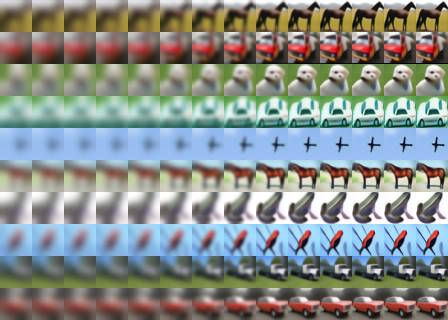}
    \caption{Progressive deblurring of selected blurred MNIST and CIFAR-10 images.}
    \label{fig:blur_prog_1}
\end{figure}

\begin{figure}
    \centering
    \includegraphics[width=\textwidth]{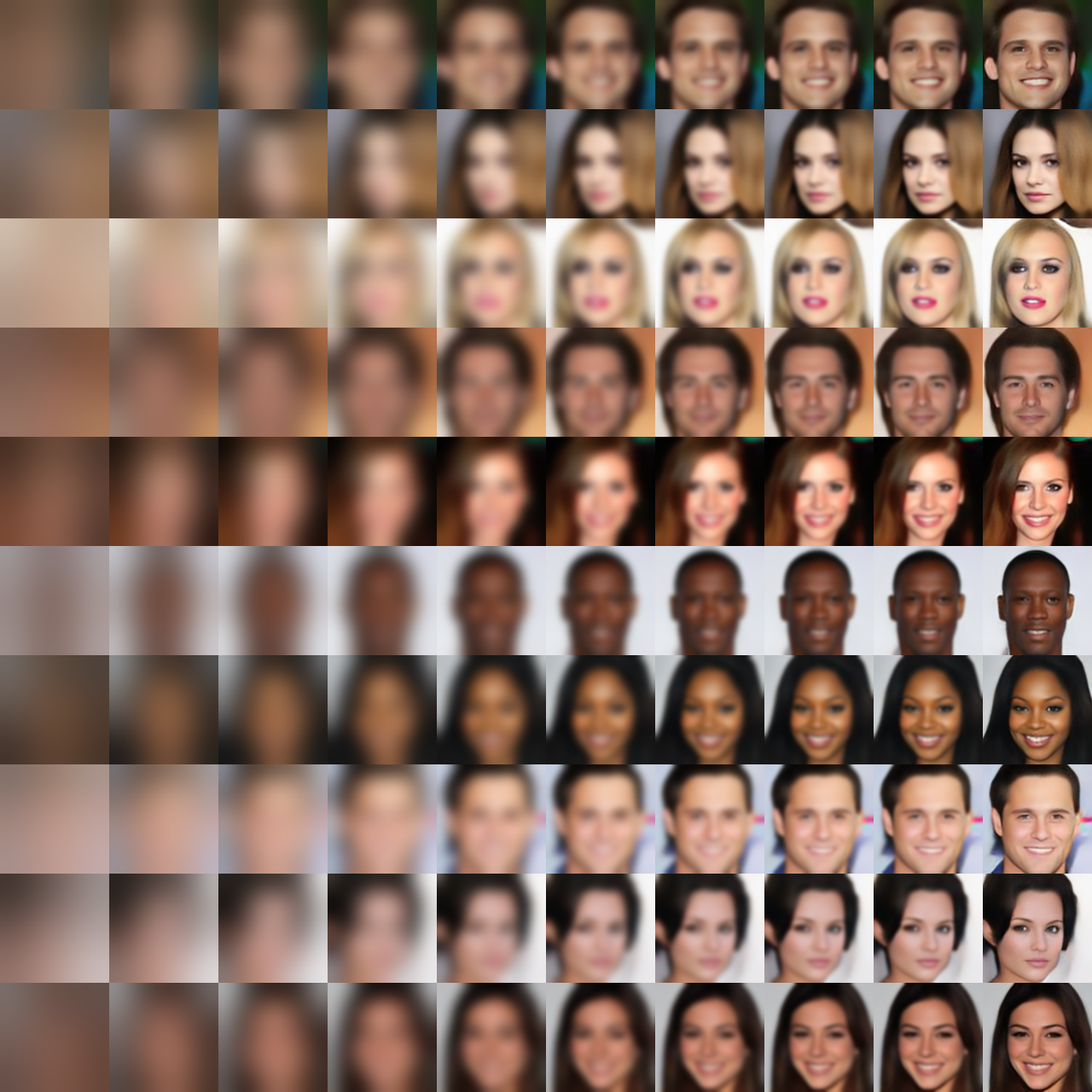}
    \caption{Progressive deblurring of selected blurred CelebA images.}
    \label{fig:blur_prog_2}
\end{figure}

\begin{figure}
    \centering
    \includegraphics[width=\textwidth]{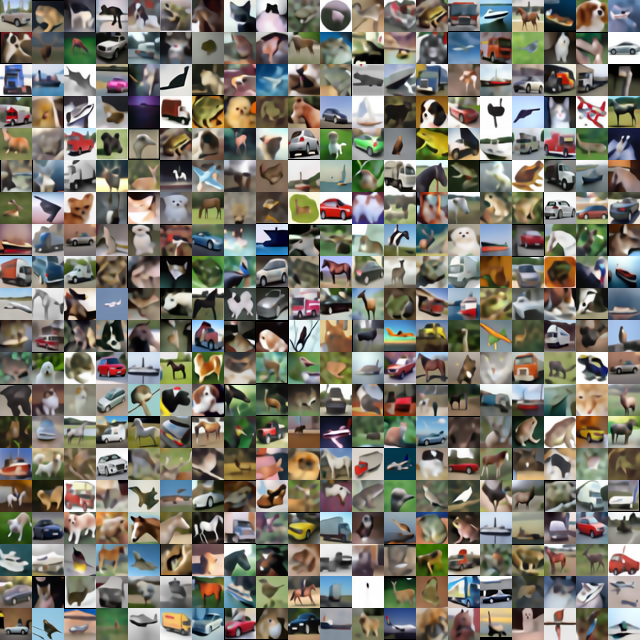}
    \caption{Deblurred Cifar10 images}
    \label{fig:cifar10_all_deblur}
\end{figure}

Figure \ref{fig:blur_revert_app} shows an additional nine images for each of MNIST, CIFAR-10 and CelebA. Figures \ref{fig:blur_prog_1} and \ref{fig:blur_prog_2} show the iterative sampling process using a deblurring model for ten example images from each dataset. We further show 400 random images to demonstrate the qualitative results in the Figure \ref{fig:cifar10_all_deblur}.

\subsection{Inpainting}

\begin{figure}[b]
    \centering
    \includegraphics[width=\textwidth]{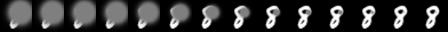}
    \includegraphics[width=\textwidth]{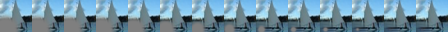}
    \includegraphics[width=\textwidth]{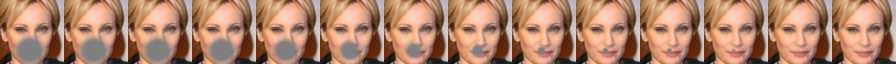}
    \caption{Progressive inpainting of selected masked MNIST, CIFAR-10, and CelebA images.}
    \label{fig:progressive_inpaint}
\end{figure}

For the inpainting transformation, models were trained on different datasets with 60,000 gradient steps. The models were trained using Adam \citep{kingma2014adam} optimizer with learning rate $2\times10^{-5}$. We use batch size 64, and the gradients are accumulated after every 2 steps. The final model is an Exponential Moving Average of the trained model with decay rate 0.995. This EMA model is updated after every 10 gradient steps. For all our inpainting experiments we use a randomized Gaussian mask and $T = 50$ with $\beta_1 = 1$ and $\beta_{i+1} = \beta_{i} + 0.1$. 

To avoid potential leakage of information due to floating point computation of the Gaussian mask, we discretize the masked image before passing it through the inpainting model. This was done by rounding all pixel values to the eight most significant digits.

\begin{figure}[h]
    \centering
    \begin{tabular}{cccc}
        Degraded & Direct & Alg. & Original \\
        \includegraphics[width=0.22\textwidth] {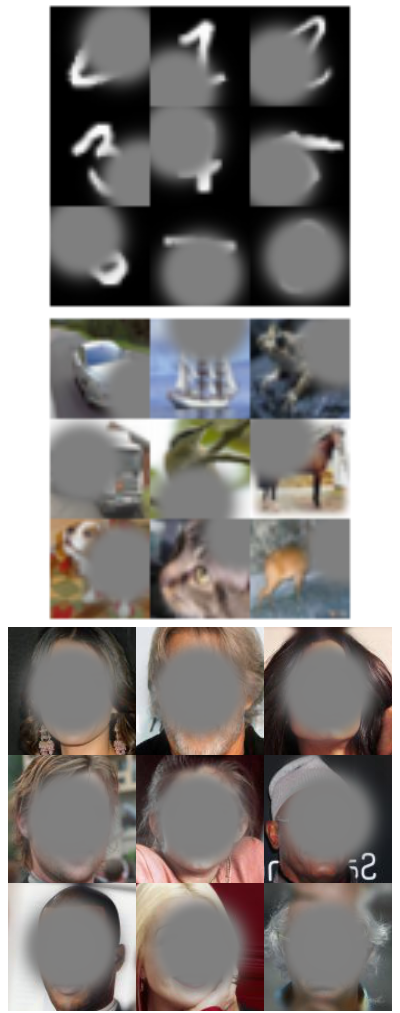} &
        \includegraphics[width=0.22\textwidth] {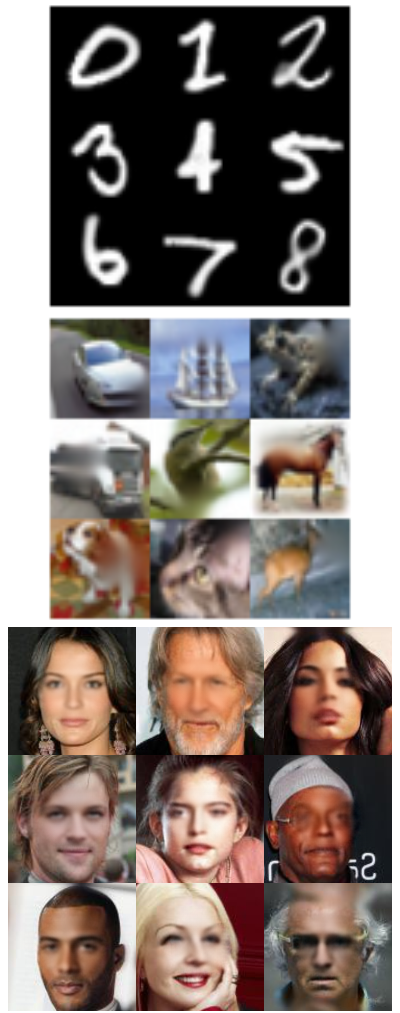} &
        \includegraphics[width=0.22\textwidth] {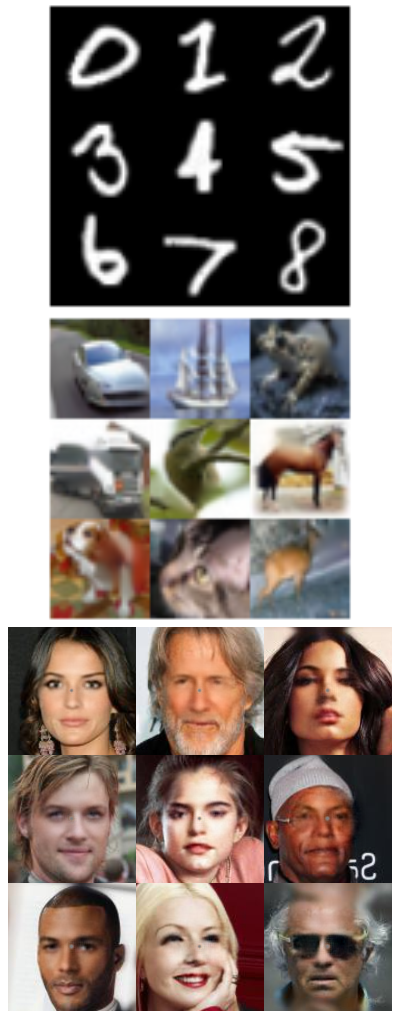} &
        \includegraphics[width=0.22\textwidth] {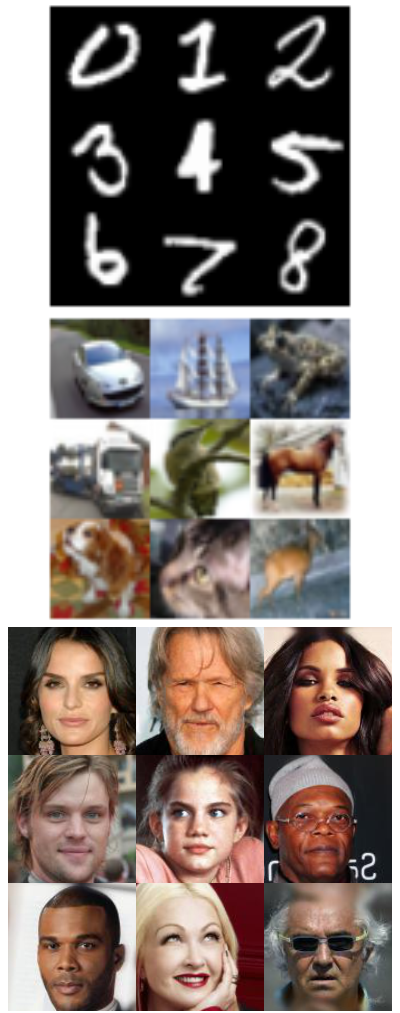}
    \end{tabular}
    \vspace{-4mm}
    \caption{Additional examples from inpainting models trained on the MNIST, CIFAR-10, and CelebA datasets. \textbf{Left to right:} degraded inputs $D(x_0, T)$ , direct reconstruction $R(D(x_0, T))$, sampled reconstruction with Algorithm \ref{alg:sa2}, and original image.
    }
    \label{fig:inpaint_revert_app}
\end{figure}

Figure \ref{fig:inpaint_revert_app} shows nine additional inpainting examples on each of the MNIST, CIFAR-10, and CelebA datasets. Figure \ref{fig:progressive_inpaint} demonstrates an example of the iterative sampling process of an inpainting model for one image in each dataset.

\subsection{Super-Resolution}

We train the super-resolution model per Section \ref{sec:training} for 700,000 iterations. We use the Adam \citep{kingma2014adam} optimizer with learning rate $2\times10^{-5}$. The batch size is 32, and we accumulate the gradients every 2 steps. Our final model is an Exponential Moving Average of the trained model with decay rate 0.995. We update the EMA model every 10 gradient steps. 

The number of time-steps depends on the size of the input image and the final image. For MNIST and for CIFAR10, the number of time steps is 3, as it takes three steps of halving the resolution to reduce the initial image down to $4 \times 4$. For CelebA, the number of time steps is 6 to reduce the initial image down to $2 \times 2$. For CIFAR10, we apply random crop and random horizontal flip for regularization. 

\begin{figure}[h]
    \centering
    \includegraphics[width=0.39\textwidth]{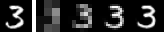}
    \includegraphics[width=0.39\textwidth]{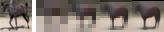}
    \includegraphics[width=0.64\textwidth]{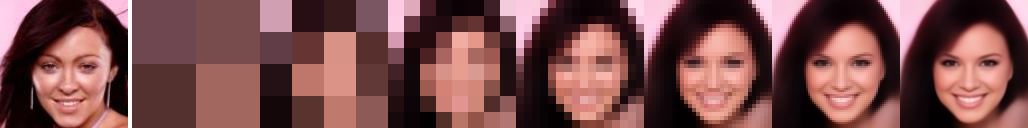}
    \caption{Progressive upsampling of selected downsampled MNIST, CIFAR-10, and CelebA images. The original image is at the left for each of these progressive upsamplings.}
    \label{fig:resolution_transform}
\end{figure}

\begin{figure}[h]
    \centering
    \begin{tabular}{cccc}
        Degraded & Direct & Alg. & Original \\
        \includegraphics[width=0.22\textwidth] {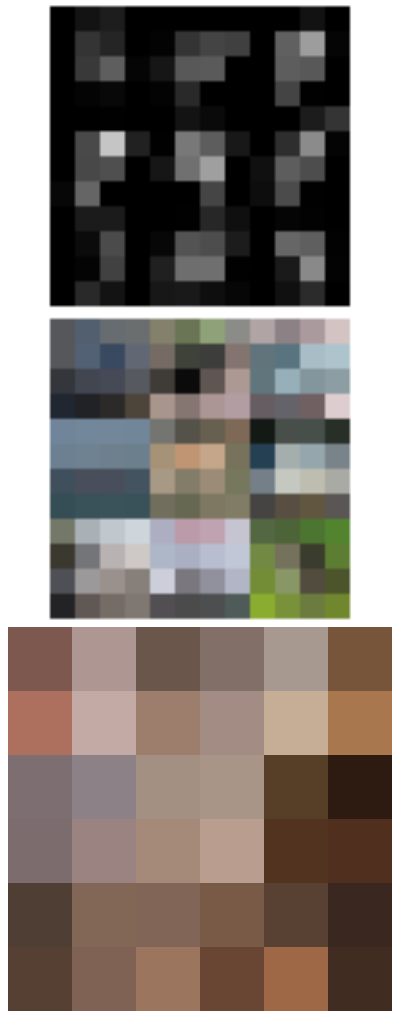} &
        \includegraphics[width=0.22\textwidth] {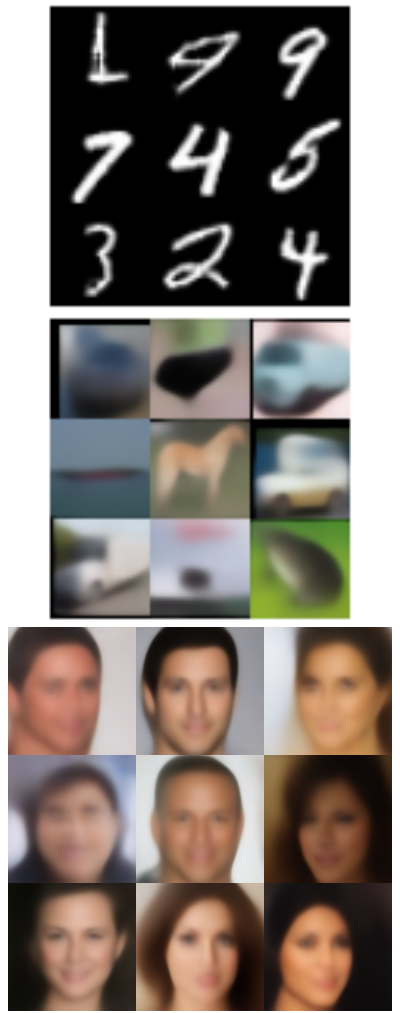} &
        \includegraphics[width=0.22\textwidth] {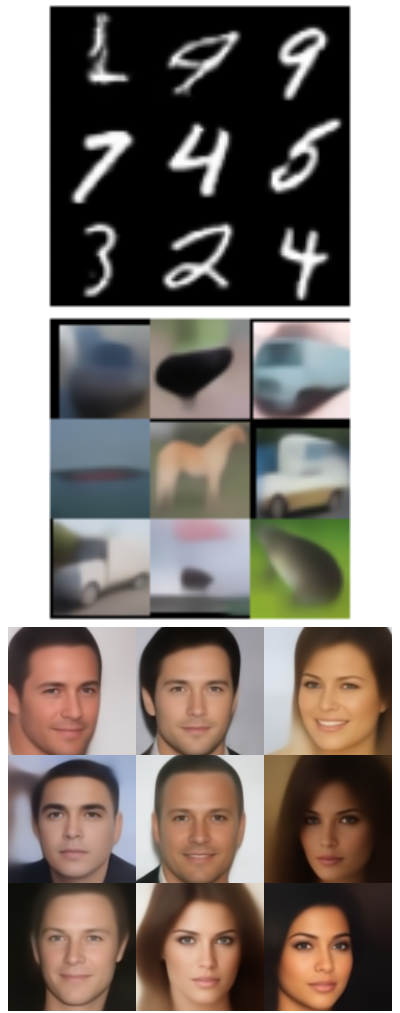} &
        \includegraphics[width=0.22\textwidth] {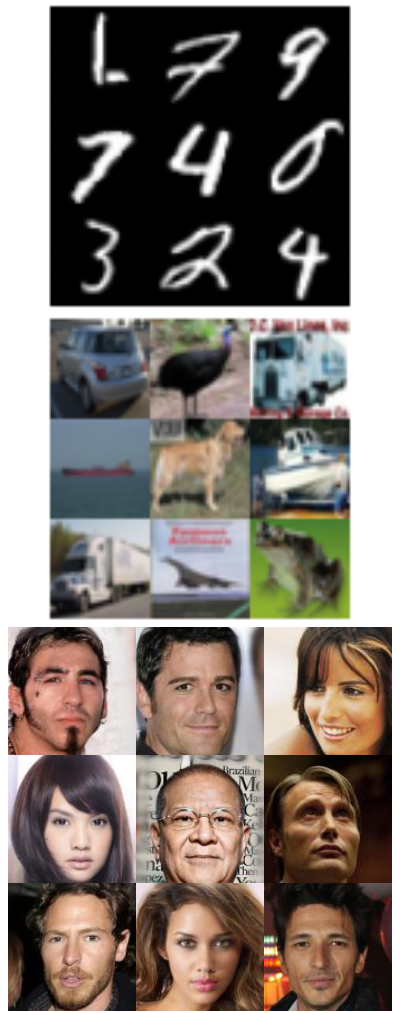}
    \end{tabular}
    \vspace{-4mm}
    \caption{Additional examples from super-resolution models trained on the MNIST, CIFAR-10, and CelebA datasets. \textbf{Left to right:} degraded inputs $D(x_0, T)$ , direct reconstruction $R(D(x_0, T))$, sampled reconstruction with Algorithm \ref{alg:sa2}, and original image.
    }
    \label{fig:resolution_revert_app}
\end{figure}

Figure \ref{fig:resolution_revert_app} shows an additional nine super-resolution examples on each of the MNIST, CIFAR-10, and CelebA datasets. Figure \ref{fig:resolution_transform} shows one example of the progressive increase in resolution achieved with the sampling process using a super-resolution model for each dataset.

\subsection{Colorization}

Here we provide results for the additional task of colorization. Starting with the original RGB-image $x_0$, we realize colorization by iteratively desaturating for $T$ steps until the final image $x_T$ is a fully gray-scale image.
We use a series of three-channel $1\times 1$ convolution filters $\mathbf{z}(\alpha) = \{z^1(\alpha), z^2(\alpha), z^3(\alpha)\}$ with the form 
\[
z^1(\alpha) = \alpha \begin{pmatrix} \frac{1}{3} \frac{1}{3} \frac{1}{3} \end{pmatrix} +  (1 - \alpha) \begin{pmatrix} 1\; 0\; 0 \end{pmatrix}
\]
\[
z^2(\alpha) = \alpha \begin{pmatrix} \frac{1}{3} \frac{1}{3} \frac{1}{3} \end{pmatrix} +  (1 - \alpha) \begin{pmatrix} 0\; 1\; 0 \end{pmatrix}
\]
\[
z^3(\alpha) = \alpha \begin{pmatrix} \frac{1}{3} \frac{1}{3} \frac{1}{3} \end{pmatrix} +  (1 - \alpha) \begin{pmatrix} 0\; 0\; 1 \end{pmatrix}
\]
and obtain $D(x, t) = \mathbf{z}(\alpha_t) * x$ via a schedule defined as $\alpha_1, \hdots, \alpha_t$ for each respective step.
Notice that a gray image is obtained when $x_T = \mathbf{z}(1) * x_0$.

We can tune the ratio $\alpha_t$  to control the amount of information removed in each step.
For our experiment, we schedule the ratio such that for every $t$ we have
\[
x_t =  \mathbf{z}(\alpha_t)* \hdots * \mathbf{z}(\alpha_1) * x_0 = \mathbf{z}(\frac{t}{T}) * x_0.
\]
This schedule ensures that color information lost between steps is smaller in earlier stage of the diffusion and becomes larger as $t$ increases. 

We train the models on different datasets for 700,000 gradient steps. We use Adam \citep{kingma2014adam} optimizer with learning rate $2\times10^{-5}$. We use batch size 32, and we accumulate the gradients every $2$ steps. Our final model is an exponential moving average of the trained model with decay rate 0.995. We update the EMA model every 10 gradient steps. For CIFAR-10 we use $T = 50$ and for CelebA we use $T = 20$.

\begin{figure}[h]
    \centering
    \begin{tabular}{cccc}
        Degraded & Direct & Alg. & Original \\
        \includegraphics[width=0.22\textwidth] {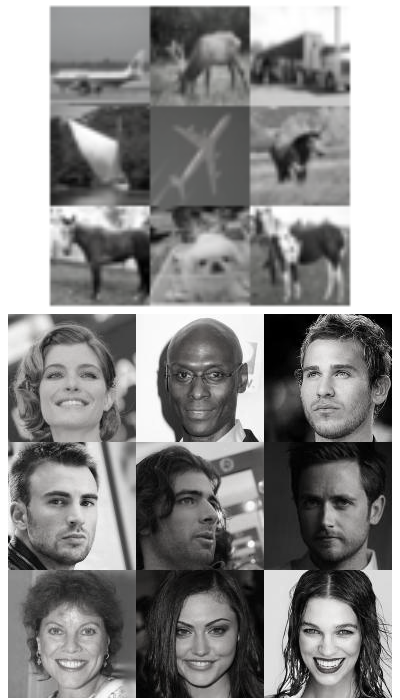} &
        \includegraphics[width=0.22\textwidth] {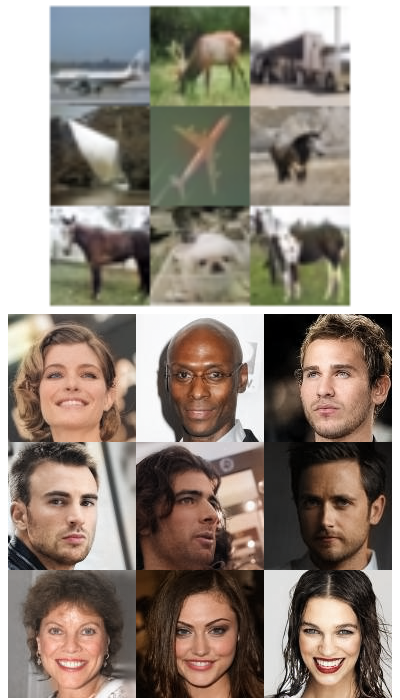} &
        \includegraphics[width=0.22\textwidth] {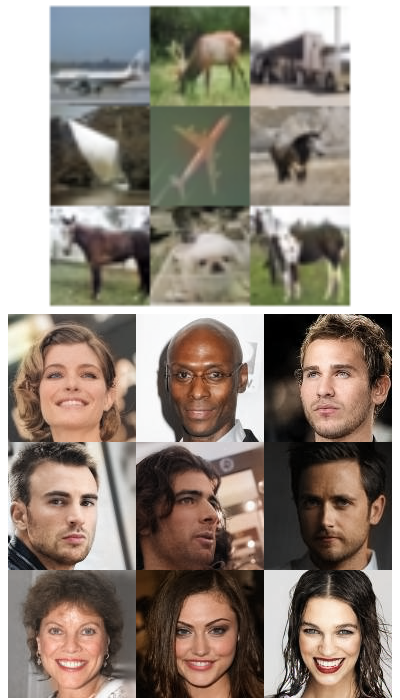} &
        \includegraphics[width=0.22\textwidth] {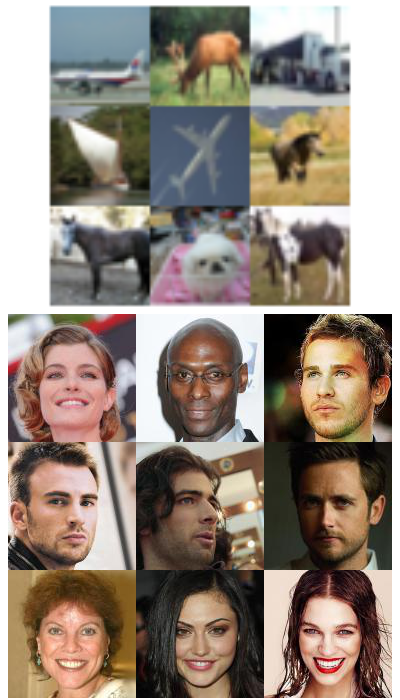}
    \end{tabular}
    \vspace{-4mm}
    \caption{Recolorization models trained on the CIFAR-10 and CelebA datasets. \textbf{Left to right:} degraded inputs $D(x_0, T)$ , direct reconstruction $R(D(x_0, T))$, sampled reconstruction with Algorithm \ref{alg:sa2}, and original image.
    }
    \label{fig:recolorize_app}
\end{figure}

\begin{table}[b]
\centering
\footnotesize
\caption{Quantitative metrics for quality of image reconstruction using recolorization models for all three channel datasets.}
\label{tab:decolorize_revert}
\vspace{0.1in}
\begin{tabular}{c|ccc|ccc}
\toprule
 & & \textbf{Degraded Image} & & & \textbf{Reconstruction} &\\
Dataset & FID &  SSIM &  RMSE & FID &  SSIM &  RMSE\\
\midrule
CIFAR-10           & 97.39 & 0.937 & 0.078 & 45.74 & 0.942 & 0.069 \\
CelebA             & 41.20 & 0.942 & 0.089 & 17.50 & 0.973 & 0.042 \\ 
\bottomrule
\end{tabular}
\end{table}

We illustrate our recolorization results in Figure \ref{fig:recolorize_app}. We present testing examples, as well as their grey scale images, from all the datasets, and compare the recolorization results with the original images. The recolored images feature correct color separation between different regions, and feature various and yet semantically correct colorization of objects. Our sampling technique still yields minor differences in comparison to the direct reconstruction, although the change is not visually apparent. We attribute this to the shape restriction of colorization task, as human perception is rather insensitive to minor color change. We also provide quantitative measurement for the effectiveness of our recolorization results in terms of different similarity metrics, and summarize the results in Table~\ref{tab:decolorize_revert}.

\subsection{Image Snow}

Here we provide results for the additional task of snowification, which is a direct adaptation of the \href{https://github.com/hendrycks/robustness}{offical implementation} of ImageNet-C snowification process~\citep{HendrycksD19imagenetc}. 
To determine the snow pattern of a given image $x_0 \in \mathbb{R}^{C\times H \times W}$, we first construct a seed matrix $S_A \in \mathbb{R}^{H\times W}$ where each entry is sampled from a Gaussian distribution $N(\mu,\sigma)$. The upper-left corner of $S_A$ is then zoomed into another matrix $S_B \in \mathbb{R}^{H\times W}$ with spline interpolation. Next, we create a new matrix $S_C$ by filtering each value of $S_B$ with a given threshold $c_1$ as
\[
S_C[i][j] = \begin{cases}
  0,         & S_B[i][j] \leq c_1 \\
  S_B[i][j], & S_B[i][j] > c_1
\end{cases}
\]
and clip each entry of $S_C$ into the range $[0,1]$.
We then convolve $S_C$ using a motion blur kernel with standard deviation $c_2$ to create the snow pattern $S$ and its up-side-down rotation $S'$. The direction of the motional blur kernel is randomly chosen as either vertical or horizontal. The final snow image is created by again clipping each value of $x_0 + S + S'$ into the range $[0,1]$. For simplicity, we abstract the process as a function $h(x_0, S_A, c_0, c_1)$.

\begin{figure}[h]
    \centering
    \begin{tabular}{cccc}
        Degraded & Direct & Alg. & Original \\
        \includegraphics[width=0.22\textwidth] {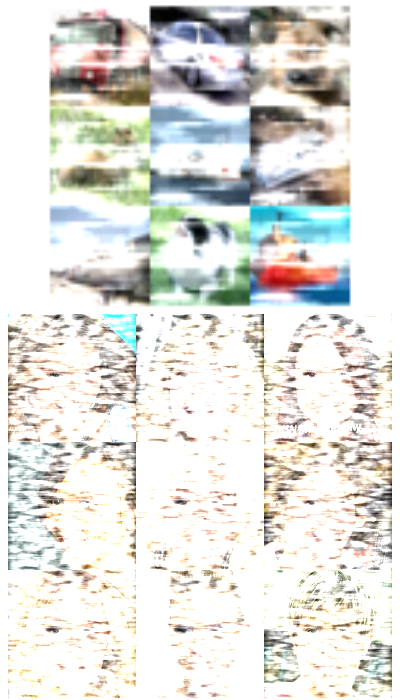} &
        \includegraphics[width=0.22\textwidth] {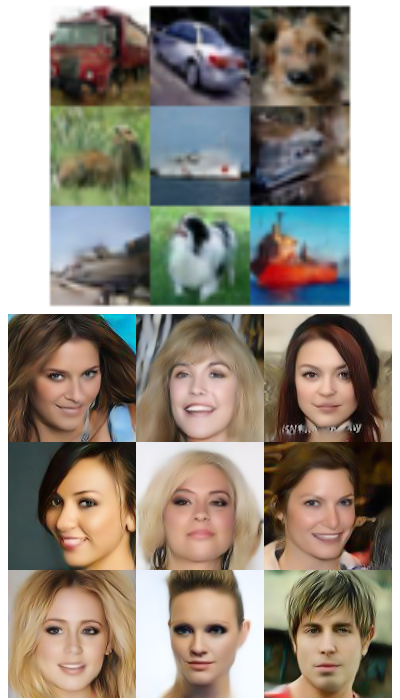} &
        \includegraphics[width=0.22\textwidth] {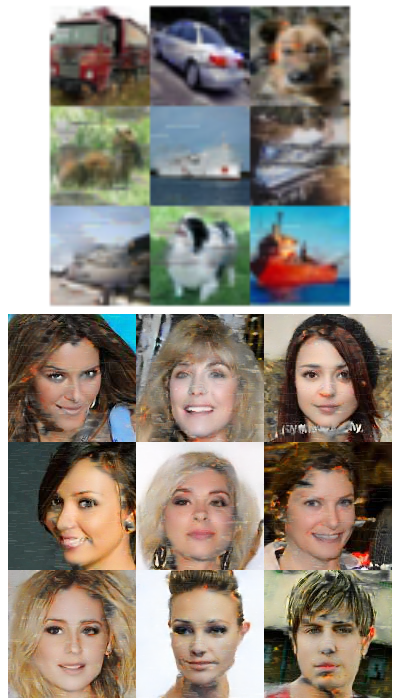} &
        \includegraphics[width=0.22\textwidth] {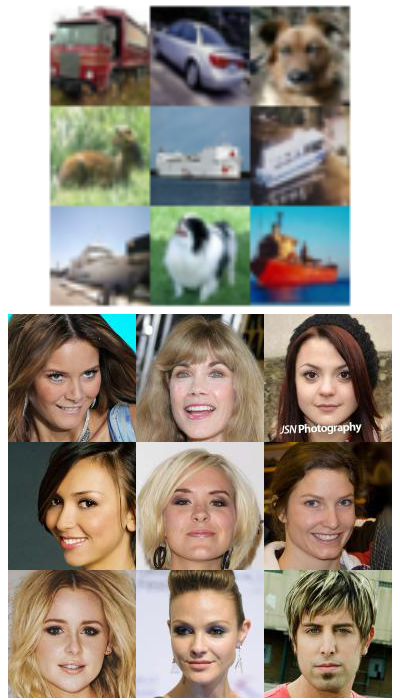}
    \end{tabular}
    \vspace{-4mm}
    \caption{Additional examples from \textit{Desnowification} models trained on the CIFAR-10 and CelebA datasets. \textbf{Left to right:} degraded inputs $D(x_0, T)$ , direct reconstruction $R(D(x_0, T))$, sampled reconstruction with Algorithm
    \ref{alg:sa2}, and original image.
    }
    \label{fig:desnowify_app}
\end{figure}

To create a series of $T$ images with increasing snowification, we linearly interpolate $c_0$ and $c_1$ between $[c_0^{\text{start}}, c_0^{\text{end}}]$ and $[c_1^{\text{start}}, c_1^{\text{end}}]$ respectively, to create $c_0(t)$ and $c_1(t)$, $t = 1, \hdots, T$. Then for each $x_0$, a seed matrix $S_x$ is sampled, the motion blur direction is randomized, and we construct each related $x_t$ by $x_t = h(x_0, S_x, c_0(t), c_1(t))$. Visually, $c_0(t)$ dictates the severity of the snow, while $c_1(t)$ determines how ``windy" the snowified image seems.

For both CIFAR-10 and Celeb-A, we use the same Gaussian distribution with parameters $\mu = 0.55$ and $\sigma = 0.3$ to generate the seed matrix.
For CIFAR-10, we choose $c_0^{\text{start}} = 1.15$, $c_0^{\text{end}} = 0.7$, $c_1^{\text{start}} = 0.05$ and $c_1^{\text{end}} = 16$, 
which generates a visually lighter snow.
For Celeb-A, we choose $c_0^{\text{start}} = 1.15$, $c_0^{\text{end}} = 0.55$, $c_1^{\text{start}} = 0.05$ and $c_1^{\text{end}} = 20$, 
which generates a visually heavier snow.

We train the models on different datasets for 700,000 gradient steps. We use Adam \citep{kingma2014adam} optimizer with learning rate $2\times10^{-5}$. We use batch size 32, and we accumulate the gradients every $2$ steps. Our final model is an exponential moving average of the trained model with decay rate 0.995. We update the EMA model every 10 gradient steps. For CIFAR-10 we use $T = 200$ and for CelebA we use $T = 200$. We note that the seed matrix is resampled for each individual training batch, and hence the snow pattern varies across the training stage.

\subsection{Generation using noise : Further Details}
\label{appendix:app_gen_noise}
Here we will discuss in further detail on the similarity between the sampling method proposed in Algorithm \ref{alg:sa2} and the deterministic sampling in DDIM \citep{DDIM_Song2021}.
Given the image $x_t$ at step $t$, we have the restored clean image $\hat{x_0}$ from the diffusion model. Hence given the estimated $\hat{x_0}$ and $x_t$, we can estimate the noise $z(x_t, t)$ (or $\hat{z}$) as
\begin{equation*}
    z(x_t, t) =  \frac{x_t - \sqrt{\alpha_t} \hat{x_0}}{\sqrt{1-\alpha_t}},
\end{equation*}
Thus, the $D(\hat{x_0}, t)$ and $D(\hat{x_0}, t-1)$ can be written as
\begin{equation*}
    D(\hat{x_0}, t) = \sqrt{\alpha_t}\hat{x_0} + \sqrt{1-\alpha_t}\hat{z},
\end{equation*}
\begin{equation*}
    D(\hat{x_0}, t-1) = \sqrt{\alpha_{t-1}}\hat{x_0} + \sqrt{1-\alpha_{t-1}}\hat{z},
\end{equation*}
using which the sampling process in Algorithm \ref{alg:sa2} to estimate $x_{t-1}$ can be written as, 

\begin{align}
x_{t-1} &= x_t - D(\hat{x_0}, t) + D(\hat{x_0}, t-1) \nonumber\\
&=x_t - (\sqrt{\alpha_t}\hat{x_0} + \sqrt{1-\alpha_t}\hat{z}) + (\sqrt{\alpha_{t-1}}\hat{x_0} + \sqrt{1-\alpha_{t-1}}\hat{z})\nonumber \\
&=\sqrt{\alpha_{t-1}}\hat{x_0} + \sqrt{1-\alpha_{t-1}}\hat{z}\nonumber \\
\end{align}
which is same as the sampling method as described in \citep{DDIM_Song2021}.

\subsection{Generation using blur transformation: Further Details}
\label{appendix:a7}

\begin{figure}[h]
    \centering
    \includegraphics[width=\textwidth]{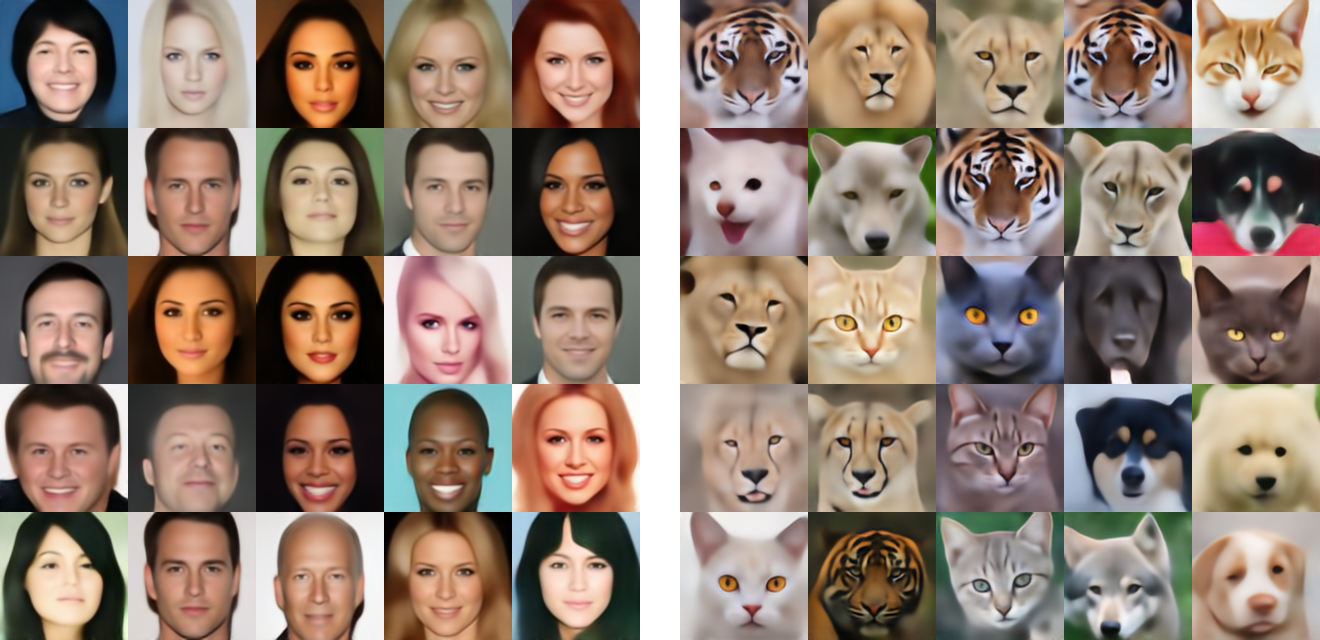}
    \caption{Examples of generated samples from $128 \times 128$ CelebA and AFHQ datasets using Method 2 with perfect symmetry.}
    \label{fig:gen_no_noise}
\end{figure}

The Figure \ref{fig:gen_no_noise}, shows the generation without breaking any symmetry within each channel are quite promising as well.

\begin{figure}[h]
    \centering
    \includegraphics[width=0.9\textwidth]{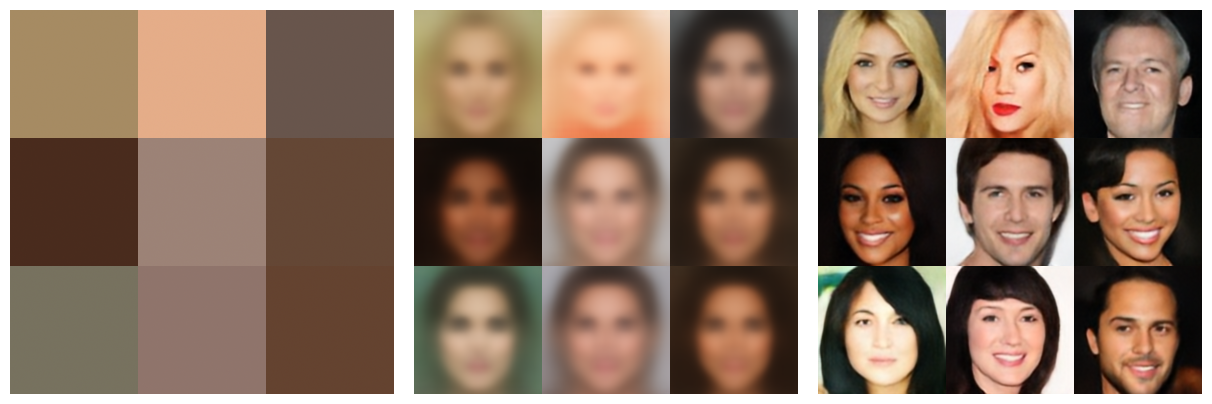}
    \includegraphics[width=0.9\textwidth]{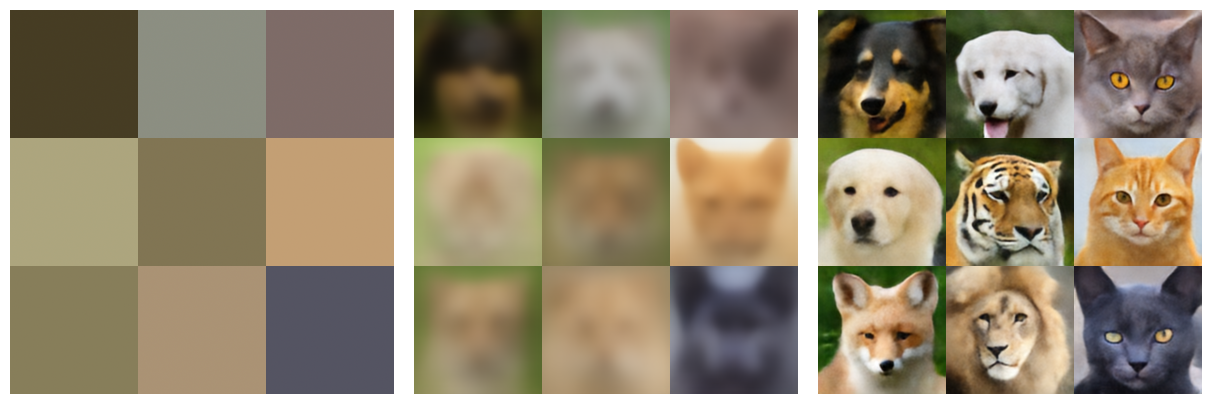}
    \caption{Comparison of direct reconstruction with sampling using Algorithm \ref{alg:sa2} for generation with Method 2 and broken symmetry. Left-hand column is the initial cold images generated using the simple Gaussian model. Middle column has images generated in one step (\textit{i.e.} direct reconstruction). Right-hand column are the images sampled with Algorithm \ref{alg:sa2}. We present results for both CelebA (top) and AFHQ (bottom) with resolution $128 \times 128$.}
    \label{fig:gen_compare_direct}
\end{figure}

\begin{figure}[t]
    \centering
    \includegraphics[width=\textwidth]{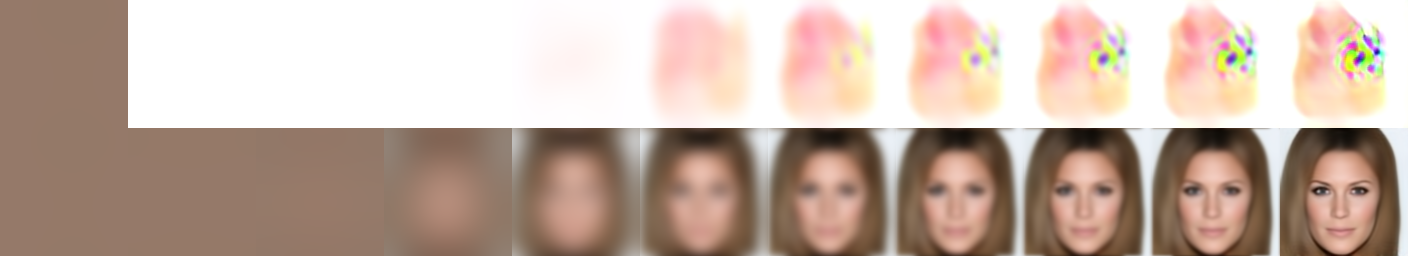}
    \caption{Comparison of Algorithm \ref{alg:sa1} (top row) and Algorithm \ref{alg:sa2} (bottom row) for generation with Method 2 and broken symmetry on $128 \times 128$ CelebA dataset. We demonstrate that Algorithm \ref{alg:sa1} fails completely to generate a new image.}
    \label{fig:gen_sampling_diff}
\end{figure}

\textbf{Necessity of Algorithm \ref{alg:sa2}:} In the case of unconditional generation, we observe a marked superiority in quality of the sampled reconstruction using Algorithm \ref{alg:sa2} over any other method considered. For example, in the broken symmetry case, the FID of the directly reconstructed images is 257.69 for CelebA and 214.24 for AFHQ, which are far worse than the scores of 49.45 and 54.68 from Table \ref{tab:fid_gen_method_2}. In Figure \ref{fig:gen_compare_direct}, we also give a qualitative comparison of this difference. We can also clearly see from Figure \ref{fig:gen_sampling_diff} that Algorithm \ref{alg:sa1}, the method used in \citet{scoreSDE_2021Song} and \citet{DDPM_Ho2020}, completely fails to produce an image close to the target data distribution. 

\end{document}